\documentclass{article} % For LaTeX2e
\usepackage{iclr2026_conference,times}

\usepackage{amsmath,amsfonts,bm}

\def\eqref#1{equation~\ref{#1}}
\def\1{\bm{1}}

\DeclareMathAlphabet{\mathsfit}{\encodingdefault}{\sfdefault}{m}{sl}
\SetMathAlphabet{\mathsfit}{bold}{\encodingdefault}{\sfdefault}{bx}{n}

\DeclareMathOperator*{\argmax}{arg\,max}

\usepackage{hyperref}
\usepackage{url}

\usepackage{amsmath}
\usepackage{amsfonts}
\usepackage{amssymb}
\usepackage{subfigure}
\usepackage{cases}
\usepackage{dsfont}
\usepackage{bm}
\usepackage{bbm}
\usepackage{graphicx}
\usepackage{color}
\usepackage{multicol}
\usepackage{multirow}
\usepackage{wrapfig,lipsum,booktabs}
\usepackage[ruled,vlined,linesnumbered]{algorithm2e}
\usepackage{pgfplots}
\pgfplotsset{compat=1.12} 
\usepackage{tikz}
\usepackage{xcolor}
\usepackage{colortbl}
\usepackage{xspace}
\usepackage{makecell}
\usepackage{soul}
\usepackage{subcaption}
\usetikzlibrary{calc}
\usepgfplotslibrary{groupplots}
\usetikzlibrary{angles,quotes} 
\usetikzlibrary{shapes,arrows}
\usetikzlibrary{matrix}
\usetikzlibrary{patterns}
\usepackage{tikz-3dplot}
\usepackage{hyperref}
\usepackage{cleveref}
\usepackage{paralist}
\usepackage{cancel}
\usepackage{tabu}
\usepackage{rotating}
\usepackage{etoolbox}
\usepackage{adjustbox}
\usepackage{enumerate}
\usepackage{enumitem}
\setitemize{itemsep=0pt,topsep=0pt,parsep=0pt,partopsep=0pt}
\setenumerate{itemsep=0pt,topsep=0pt,parsep=0pt,partopsep=0pt}
\usepackage{pifont}
\usepackage{cancel}
\usepackage{lipsum}
\usepackage{listings,lstautogobble}
\usepackage{fancyvrb}
\usepackage{fvextra}
\usepackage{caption}
\usepackage{arydshln}
\usepackage{float}
\usepackage{color} 
\usepackage{colortbl}
\usepackage{pmboxdraw} % ├  └
\usepackage{wasysym}
\usepackage{csquotes}
\usepackage{mfirstuc}
\usepackage{algorithmic}

\usepackage[most,skins,theorems]{tcolorbox}

\definecolor{summfr}{RGB}{220, 108, 110} % Frame color
\definecolor{bg}{RGB}{224, 214, 250} % Background color

\definecolor{quizfr}{RGB}{176, 208, 160} % Frame color

\definecolor{examfr}{RGB}{247,179,1} % Frame color

\definecolor{judgefr}{RGB}{104,164,241} % Frame color

\newtcbtheorem{SummGen}{Summary Generators}{colback=bg!25,colframe=summfr, boxrule=0.5pt}{th}

\newtcbtheorem{QuizGen}{Quiz Generators}{colback=bg!25,colframe=quizfr, boxrule=0.5pt
}{th}

\newtcbtheorem{SummReview}{Summary Reviewers}{colback=bg!25,colframe=summfr, boxrule=0.5pt}{th}

\newtcbtheorem{QuizReview}{Quiz Reviewers}{colback=bg!25,colframe=quizfr, boxrule=0.5pt}{th}

\newtcbtheorem{Examinee}{Examinee}{colback=bg!25,colframe=examfr, boxrule=0.5pt}{th}

\newtcbtheorem{LLM-as-a-Judge}{LLM-as-a-Judge}{colback=bg!25,colframe=judgefr, boxrule=0.5pt}{th}

\newcommand{\model}[1]{\textsc{#1}\xspace}
\newcommand{\ours}{\model{SummQ}}
\newcommand{\ourssolo}{\model{SummQ\textsubscript{solo}}}
\newcommand{\oursensemble}{\model{SummQ\textsubscript{combo}}}
\newcommand{\oursensemblerephrase}{\model{SummQ\textsubscript{comboR}}}

\newcommand{\textrank}{\model{TextRank}}
\newcommand{\longtfive}{\model{LongT5}}
\newcommand{\unlim}{\model{U.former}}
\newcommand{\sled}{\model{SLED}}
\newcommand{\cached}{\model{CachED}}
\newcommand{\gptfouro}{\model{GPT-4o}}
\newcommand{\gptfouromini}{\model{GPT-4o-mini}}
\newcommand{\gptfourone}{\model{GPT-4.1}}
\newcommand{\gptfive}{\model{GPT-5}}
\newcommand{\gptothree}{\model{o3}}
\newcommand{\dpsk}{\model{DeepSeek-R1}}
\newcommand{\qwen}{\model{Qwen3-32B}}
\newcommand{\hmsr}{\model{HM-SR}}
\newcommand{\cmllm}{\model{C.MultiLLM}}

\newcommand{\dataset}[1]{\texttt{#1}\xspace}
\newcommand{\mensa}{\dataset{MENSA}}
\newcommand{\booksum}{\dataset{BookSum}}
\newcommand{\govreport}{\dataset{GovReport}}

\newcommand{\rougeone}{R-1\xspace}
\newcommand{\rougetwo}{R-2\xspace}
\newcommand{\rougel}{R-L\xspace}

\newcommand{\bsp}{BS\textsubscript{$P$}\xspace}
\newcommand{\bsr}{BS\textsubscript{$R$}\xspace}
\newcommand{\bsf}{BS\textsubscript{$F_1$}\xspace}

\newcommand{\cmark}{\ding{51}}%

\SetCommentSty{mycommfont}

\title{Learning to Summarize by Learning to Quiz: Adversarial Agentic Collaboration for Long Document Summarization}

\iclrfinalcopy

\author{%
Weixuan Wang\textsuperscript{1$\dagger$} \quad Minghao Wu\textsuperscript{2$\dagger$} \quad Barry Haddow\textsuperscript{1} \quad Alexandra Birch\textsuperscript{1} \\[1ex]
\textsuperscript{1}School of Informatics, University of Edinburgh \\
\textsuperscript{2}Tongyi Lab, Alibaba Group \\
\texttt{\{weixuan.wang, bhaddow, a.birch\}@ed.ac.uk} \\
\texttt{minghao.wu@alibaba-inc.com}
}

\begin{document}

\maketitle

\begingroup
\renewcommand\thefootnote{}\footnotetext{
\textsuperscript{$\dagger$}~Equal contribution.
}
\endgroup

\begin{abstract}

Long document summarization remains a significant challenge for current large language models (LLMs), as existing approaches commonly struggle with information loss, factual inconsistencies, and coherence issues when processing excessively long documents. We propose \ours, a novel adversarial multi-agent framework that addresses these limitations through collaborative intelligence between specialized agents operating in two complementary domains: \textit{summarization} and \textit{quizzing}. Our approach employs summary generators and reviewers that work collaboratively to create and evaluate comprehensive summaries, while quiz generators and reviewers create comprehension questions that serve as continuous quality checks for the summarization process. This adversarial dynamic, enhanced by an examinee agent that validates whether the generated summary contains the information needed to answer the quiz questions, enables iterative refinement through multifaceted feedback mechanisms. We evaluate \ours on three widely used long document summarization benchmarks. Experimental results demonstrate that our framework significantly outperforms existing state-of-the-art methods across ROUGE and BERTScore metrics, as well as in LLM-as-a-Judge and human evaluations. Our comprehensive analyses reveal the effectiveness of the multi-agent collaboration dynamics, the influence of different agent configurations, and the impact of the quizzing mechanism. This work establishes a new approach for long document summarization that uses adversarial agentic collaboration to improve summarization quality. In addition, we release the code to foster research along this line.\footnote{\url{https://github.com/weixuan-wang123/SummQ}}

\end{abstract}

\section{Introduction}

Summarization has become increasingly critical in modern natural language processing, as organizations and individuals face an ever-growing volume of textual information that requires efficient processing and comprehension \citep{DBLP:journals/air/GambhirG17,zhao2020seal,zhang2021summ}. Prior works in summarization have primarily focused on short to medium-length documents, where models can effectively capture the essential content and generate coherent summaries \citep{see-etal-2017-get,fabbri-etal-2019-multi}. Recently, there has been an increasing interest in long document document summarization, driven by the need to process extensive texts such as research articles, legal documents, and books \citep{huang-etal-2021-efficient,kryscinski-etal-2022-booksum,saxena-keller-2024-select}.

Recent large language models (LLMs) have shown promising results for summarization tasks \citep{pu2023summarization,laban-etal-2023-summedits, keswani2024abstractive}. However, existing methods struggle with long documents, often leading to significant information loss, factual inconsistencies, and difficulty maintaining coherence across lengthy texts \citep{DBLP:journals/csur/KohJLP23}. Current approaches often fail to capture the nuanced relationships between distant parts of a document, resulting in summaries that may miss crucial information or introduce hallucinations \citep{chrysostomou-etal-2024-investigating,tang-etal-2024-tofueval,xia-etal-2024-hallucination}. Recently, multi-agent systems has demonstrated potential for improving complex reasoning tasks through collaborative interactions \citep{DBLP:conf/ijcai/GuoCWCPCW024}, yet their application to long document summarization remains underexplored \citep{fang2024multi,kim-kim-2025-nexussum}.

\begin{figure}[t]
    \centering
    \includegraphics[scale=0.42]{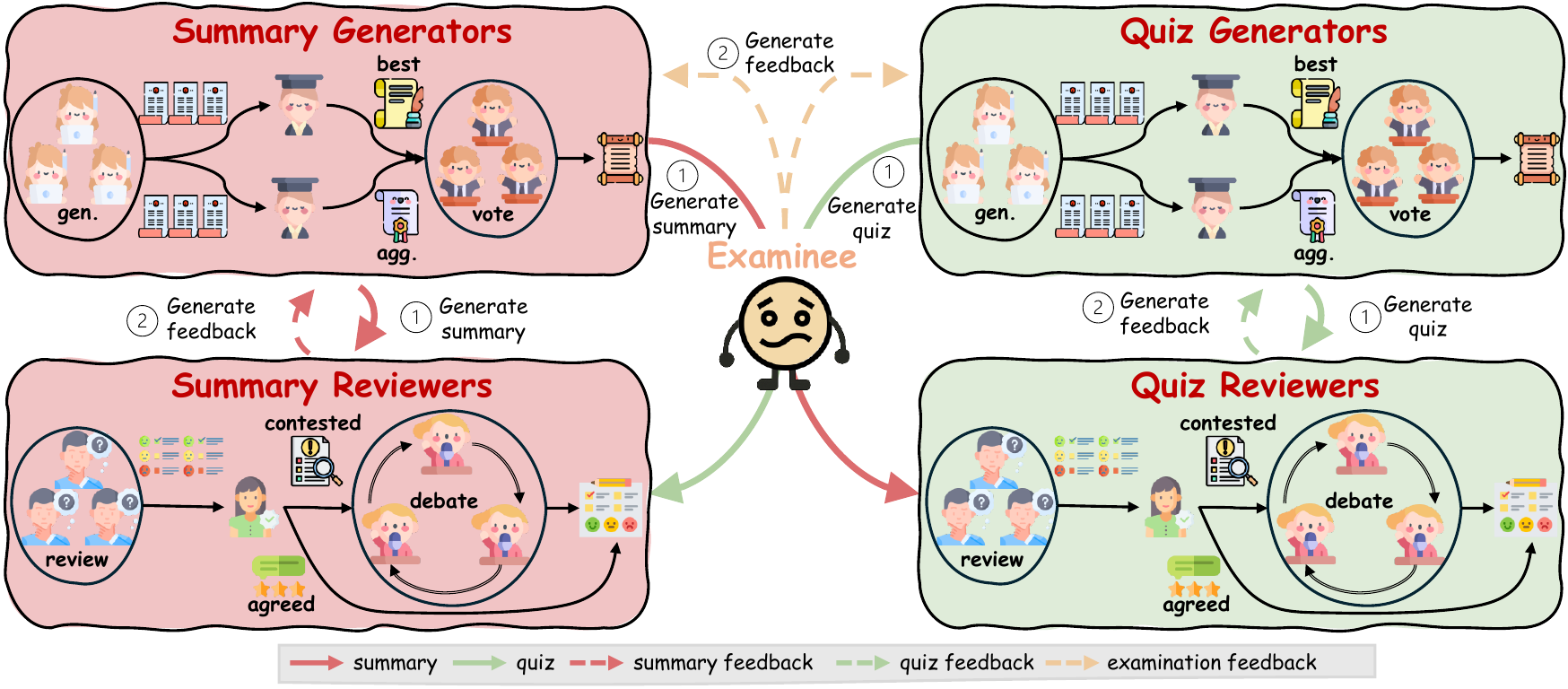}
    \caption{
    The overall framework of \ours. It consists of two tasks, summarization and quizzing, and two types of agents: generator and reviewer, resulting in four groups of agents: \textit{Summary Generators}, \textit{Quiz Generators}, \textit{Summary Reviewers}, and \textit{Quiz Reviewers}. Additionally, we include an \textit{Examinee} agent to check if quiz questions can be answered by the summary.
    }
    \label{fig:framework}
\end{figure}

To address these challenges, we propose \ours, an adversarial multi-agent framework that leverages collaborative intelligence to generate high-quality summaries for long documents. Our approach divides specialized agents into two complementary task domains: \textit{summarization} and \textit{quizzing}. Within the summarization domain, we deploy summary generators that collaboratively create comprehensive summaries through independent drafting, aggregation, and collective voting, alongside summary reviewers that rigorously evaluate content quality through independent review and structured debate mechanisms. Simultaneously, the quizzing domain employs quiz generators to create comprehension quizzes that test the completeness and accuracy of generated summaries, while quiz reviewers ensure the quality, coverage, and appropriateness of these assessments. This dual-task framework creates a natural adversarial dynamic where the quiz generation process serves as a continuous quality check for summarization. The summary aims to provide comprehensive coverage of the document, enabling the quiz questions to be answered correctly, while the quiz challenges the information coverage, factuality, and coherence of the summary. Furthermore, an examinee agent is introduced to provide additional feedback, ensuring that the quiz questions can be accurately answered using only the generated summary. Through iterative refinement guided by multifaceted feedback, \ours ensures that the final summaries are not only comprehensive and coherent but also factually accurate and verifiable.

To validate the effectiveness of \ours, we conduct extensive experiments on three long document summarization tasks including \mensa \citep{saxena-keller-2024-select}, \booksum \citep{kryscinski-etal-2022-booksum}, and \govreport \citep{huang-etal-2021-efficient}. Our results demonstrate that \ours significantly outperforms existing state-of-the-art methods in terms of ROUGE \citep{lin-2004-rouge} and BERTScore \citep{DBLP:conf/iclr/ZhangKWWA20}, as well as LLM-as-a-Judge and human evaluations. Furthermore, more in-depth analyses highlight the effectiveness of our multi-agent framework in enhancing summary quality, the coverage of the generated quizzes, and the impact of various agent configurations.

Our contributions are summarized as follows:
\begin{itemize}
    \item We introduce \ours, a novel adversarial multi-agent framework that integrates summarization and quizzing tasks to enhance the long document summarization (see \autoref{sec:method}).
    \item We conduct comprehensive experiments on three long document summarization benchmarks, demonstrating that \ours achieves state-of-the-art performance across multiple evaluation metrics and human assessments (see \autoref{sec:experiments}).
    \item We provide in-depth analyses of the multi-agent collaboration, the dynamics of the quizzing mechanism, and the impact of various agent configurations (see \autoref{sec:analysis}).
\end{itemize}
\section{Related Work}

\paragraph{Multi-Agent Systems}
Recent advances in LLMs have enabled the development of multi-agent systems that harness the strengths of multiple agents to tackle complex tasks \citep{wang2024survey,DBLP:conf/ijcai/GuoCWCPCW024,xi2025rise}. These systems typically involve agent collaboration to boost performance, as seen in multi-agent debating \citep{DBLP:conf/icml/Du00TM24,DBLP:conf/emnlp/XiongD00023,DBLP:journals/corr/abs-2310-20151,DBLP:conf/acl/TangZ0L0ZCG24} and discussion \citep{DBLP:conf/acl/ChenSB24,DBLP:conf/naacl/SahaLCBWL24} for reasoning over short texts \citep{DBLP:conf/icml/Du00TM24,DBLP:conf/acl/TangZ0L0ZCG24}, paper review \citep{DBLP:journals/corr/abs-2311-08152}, dataset synthesis \citep{wang2025odysseybench}, machine translation \citep{DBLP:journals/corr/abs-2405-11804}, and code generation \citep{DBLP:journals/corr/abs-2312-13010,DBLP:conf/coling/WangR0LBCYZYSL25}. Collaboration among agents introduces diverse perspectives and complementary skills, leading to higher-quality and more robust outputs.

\paragraph{Long document Summarization}
Long document summarization has seen various methods, including architectural optimizations (e.g., sparse attention \citep{DBLP:conf/emnlp/0006ZWX0Y21,ivgi2023efficient,bertsch2023unlimiformer}, long-context finetuning \citep{DBLP:journals/corr/abs-2004-05150,guo2021longt5}, memory augmentation \citep{cui2021sliding,saxena2025end}, window extension \citep{press2021train,DBLP:journals/corr/abs-2306-15595,DBLP:journals/ijon/SuALPBL24,DBLP:conf/acl/YenG024}) and chunking strategies (e.g., sliding window \citep{zaheer2020big,DBLP:conf/eacl/PangNKSZX23}. 
LLMs with improved long-context abilities have shifted the field toward leveraging their strong language skills for summarization \citep{goyal2022news,DBLP:conf/acl/RatnerLBRMAKSLS23,keswani2024abstractive}, but still face challenges with context limits and maintaining coherence \citep{pu2023summarization,liu2023lost}. Multi-agent systems have been explored to address these issues, enabling collaborative, more accurate summaries \citep{DBLP:journals/corr/abs-2402-11550,fang2024multi,jeong2025agent}, though many still rely on self-verification, leading to biases and missed errors.

\section{Methodology}
\label{sec:method}

In this section, we present \ours for long document summarization, as illustrated in \autoref{fig:framework}. We first introduce the overall workflow in \autoref{sec:method_overall_workflow}, and then describe the collaboration between the generator and reviewer agents in \autoref{sec:method_generator} and \autoref{sec:method_reviewer}, respectively.

\begin{algorithm}[htbp]
\caption{Overall \ours Workflow}
\label{alg:overall}
\KwIn{Document $D$; Summary Generators $\mathcal{G}_s$; Quiz Generators $\mathcal{G}_q$; Summary Reviewers $\mathcal{R}_s$; Quiz Reviewers $\mathcal{R}_q$; Examinee $\mathcal{E}$; Number of iterations $T_{\text{iter}}$}
\KwOut{Accepted summary $S^*$, accepted quiz $Q^*$}
$S^{(0)} \leftarrow \emptyset$; $Q^{(0)} \leftarrow \emptyset$ \tcp*{Initialize summary $S^{(0)}$ and quiz $Q^{(0)}$}

\For{iteration $t = 1$ \textbf{to} $T_{\text{iter}}$}{
    % Generation phase
    $S^{(t)} \leftarrow \textsc{Generate}(\mathcal{G}_s, D, S^{(t-1)})$ \tcp*{Summary Generators produce candidate summaries}
    $Q^{(t)} \leftarrow \textsc{Generate}(\mathcal{G}_q, D, Q^{(t-1)})$ \tcp*{Quiz Generators produce candidate quizzes}

    % Review phase
    $F_s^{(t)} \leftarrow \textsc{Review}(\mathcal{R}_s, S^{(t)}, Q^{(t)}, D)$ \tcp*{Summary Reviewers produce feedback on summary}
    $F_q^{(t)} \leftarrow \textsc{Review}(\mathcal{R}_q, Q^{(t)}, S^{(t)}, D)$ \tcp*{Quiz Reviewers produce feedback on quiz}

    % Answer Review
    $F_e^{(t)} \leftarrow \textsc{TakeQuiz}(\mathcal{E}, Q^{(t)}, S^{(t)})$ \tcp*{Examinee $\mathcal{E}$ takes the quiz based on the summary}

    % Aggregated feedback
    $F_s^{(t)} \leftarrow F_s^{(t)} \cup F_e^{(t)}|_{\text{summary}}$ \tcp*{Merge feedback relevant to summary}
    $F_q^{(t)} \leftarrow F_q^{(t)} \cup F_e^{(t)}|_{\text{quiz}}$ \tcp*{Merge feedback relevant to quiz}

    % Acceptance check
    \If{$F_s^{(t)} = \emptyset$ \textbf{and} $F_q^{(t)} = \emptyset$}{
        \Return $(S^{(t)}, Q^{(t)})$ \tcp*{If no issues, accept and return the summary and quiz}
    }
    % \If{${F_s^{(t)}}.\mathit{dec} = \text{ACCEPT}$ \textbf{and} ${F_q^{(t)}}.\mathit{dec} = \text{ACCEPT}$}{
    %     \Return $(S^{(t)}, Q^{(t)})$ \tcp*{If no issues, accept and return the summary and quiz}
    % }

}
\Return $(S^{(T)}, Q^{(T)})$
\end{algorithm}

\subsection{Overall Workflow}
\label{sec:method_overall_workflow}
The overall workflow of \ours, depicted in \autoref{fig:framework}, involves two primary tasks: summarization and quiz generation, supported by two types of agents: generators $\mathcal{G}$ and reviewers $\mathcal{R}$. These tasks and agents combine to form four distinct components: \textit{Summary Generators $\mathcal{G}_{s}$}, \textit{Quiz Generators $\mathcal{G}_{q}$}, \textit{Summary Reviewers $\mathcal{R}_{s}$}, and \textit{Quiz Reviewers $\mathcal{R}_{q}$}. Additionally, an \textit{Examinee} agent $\mathcal{E}$ is incorporated to validate that the quiz questions can be accurately answered using the summary.

The interaction between summarization and quiz generation creates a natural adversarial framework that continuously improves summarization quality. In this framework, the summary aims to provide comprehensive coverage that enables correct answers to quiz questions, while the quiz challenges the information coverage, factuality, and coherence of the summary. This dual-task approach ensures that both components evolve together, resulting in summaries that are not only informative but also verifiable through targeted questioning.

The iterative workflow of \ours, as detailed in \autoref{alg:overall}, operates through a systematic process of generation, reviewing, and refinement. Beginning with an input document $D$, the system initializes empty summary and quiz states and enters an iterative loop for up to $T_{iter}$ iterations. In each iteration $t$, the process unfolds in four key stages. First, the Summary Generators $\mathcal{G}_s$ produce a candidate summary $S^{(t)}$ based on the document and any previous summary, while Quiz Generators $\mathcal{G}_q$ simultaneously generate a candidate quiz $Q^{(t)}$. Second, the reviewing phase begins with Summary Reviewers $\mathcal{R}_s$ evaluating the generated summary against both the quiz and original document to produce feedback $F_s^{(t)}$, and Quiz Reviewers $\mathcal{R}_q$ assessing the quiz quality to generate feedback $F_q^{(t)}$. Third, an Examinee agent $\mathcal{E}$ attempts to answer the quiz questions using only the generated summary, providing additional feedback $F_e^{(t)}$ that is then merged with the respective summary and quiz feedback streams. Finally, the system performs an acceptance check: if both feedback sets are empty, the current summary and quiz are accepted and returned. This iterative refinement continues until either high-quality outputs are achieved or the maximum iteration limit is reached. Note that \textsc{Generate()} and \textsc{Review()} are detailed in \autoref{sec:method_generator} and \autoref{sec:method_reviewer}, respectively.

This comprehensive workflow design ensures allows for continuous improvement based on concrete feedback, while the dual-task approach of simultaneous summary and quiz generation creates a natural consistency check that enhances the overall reliability and coherence of the final outputs.

\begin{algorithm}[htbp]

\caption{\textsc{Generate()}: Generator Collaboration}
\label{alg:generator}
\KwIn{Document $D$; Previous summary/quiz $z'$; Generator agents $\mathcal{G}=\{ g_i \}_{i=1}^n $; Aggregator agent $A_{\text{Agg}}$; Ranker agent $A_{\text{Ranker}}$}
\KwOut{Final summary/quiz $z^*$}

\BlankLine
\textbf{Phase 1: Independent Draft Generation}\;
\For{each generator agent $g_i \in \mathcal{G}$}{
    $z_i \leftarrow \textsc{Draft}(g_i, D, z')$ \tcp*{Generate independent draft summary/quiz}
}
$\mathcal{Z} = \{z_1, z_2, \ldots, z_n\}$ \tcp*{Set of all draft summaries/quizzes}

\BlankLine
\textbf{Phase 2: Draft Aggregation}\;
$z_{\text{agg}} \leftarrow \textsc{Aggregate}(A_\text{Agg}, \mathcal{Z})$ \tcp*{Aggregate drafts into an unified summary/quiz}

\BlankLine
\textbf{Phase 3: Best Draft Selection}\;
$z_{\text{best}} \leftarrow \textsc{BestSelect}(A_\text{Ranker}, \mathcal{Z})$ \tcp*{Select best individual draft}

\BlankLine
\textbf{Phase 4: Collective Voting}\;
$\mathcal{C} \leftarrow \{z_{\text{agg}}, z_{\text{best}}\}$ \tcp*{Candidate summaries/quizzes for voting}
\For{each agent $g_j \in \mathcal{G}$}{
    $\text{vote}_j \leftarrow \textsc{Prefer}(g_j, \mathcal{C}, D)$ \tcp*{Agent $g_j$ votes for preferred candidate}
}
$z^* \leftarrow \argmax_{z \in \mathcal{C}} \left|\{j : \text{vote}_j = z\}\right|$ \tcp*{Select the candidate with the most votes}

\BlankLine
\Return $z^*$
\end{algorithm}

\paragraph{Quiz Generation}
The Quiz Generators $\mathcal{G}_q$ are responsible for producing a diverse range of question types, including multiple-choice, true-false, and short-answer questions. Through collaboration among multiple Quiz Generators, the system generates 10 questions for each type as well as the corresponding answers, resulting in a total of 30 question-answer pairs per quiz.

\subsection{Generator Collaboration}
\label{sec:method_generator}

The generator collaboration in \ours is built around a multi-phase process that combines the strengths of multiple generator agents. As shown in \autoref{alg:generator}, this process unfolds in four phases:

\paragraph{Phase 1: Independent Draft Generation} The process begins with each generator agent $g_i \in \mathcal{G}$ working independently to create its own draft summary/quiz $z_i$ from the input document $D$ and any previous summary or quiz $z'$. This parallel approach naturally leads to diverse initial drafts, since different agents may emphasize various aspects of the document. We collect all these draft summaries or quizzes into a set $\mathcal{Z} = \{z_1, z_2, \ldots, z_n\}$, where $n$ denotes the total number of generator agents involved in the process.

\paragraph{Phase 2: Draft Aggregation} In the second phase, an aggregator agent $A_\text{Agg}$ combines the individual draft summaries or quizzes into a unified summary or quiz $z_{\text{agg}}$. This agent selectively combines the strengths of each draft and incorporates complementary information that individual agents may have overlooked. By drawing upon the collective knowledge across all drafts, $A_\text{Agg}$ creates a more comprehensive summary or quiz that harnesses the diverse perspectives and insights.

\paragraph{Phase 3: Best Draft Selection} Concurrently with the aggregation process, a ranker agent $A_\text{Ranker}$ evaluates each of the individual draft in $\mathcal{Z}$ to identify the highest-quality draft $z_{\text{best}}$. This parallel selection process serves as an important safeguard, ensuring that when a particular agent produces an exceptionally strong summary or quiz, it remains visible and is not overshadowed by $A_\text{Agg}$.

\paragraph{Phase 4: Collective Voting} In the final phase, we bring together the collective wisdom of all generator agents to make the ultimate decision between two candidates: the aggregated summary/quiz $z_{\text{agg}}$ and the best individual draft $z_{\text{best}}$. Each generator agent $g_j$ carefully evaluates both candidates from the set $\mathcal{C} = \{z_{\text{agg}}, z_{\text{best}}\}$ and casts their vote for the one they believe best captures the essence of the original document. The final summary/quiz is the candidate that receives the most votes.

\begin{algorithm}[htbp]
\caption{\textsc{Review()}: Reviewer Collaboration}
\label{alg:reviewer}
\KwIn{Document $D$; Summary/Quiz $z$; Reviewer agents $\mathcal{R} = \{ r_i \}^{n}_{i=1}$; Number of debate rounds $T_{\text{debate}}$}
\KwOut{Decision $\mathit{dec} \in \{\text{ACCEPT}, \text{REJECT}\}$; Issue list $\mathcal{I}$}

\textbf{Phase 1: Independent Reviewing} \;
\For{each reviewer $r_i \in \mathcal{R}$}{
    $\mathcal{A}_i \leftarrow \textsc{Annotate}(r_i, z, D)$ \tcp*{Review and annotate the summary/quiz}
}

\textbf{Phase 2: Issue Categorization} \;
$\mathcal{M} \leftarrow \left\{ a \mid a \in \bigcup_{i=1}^n \mathcal{A}_i \text{ and } \left| \{ i : a \in \mathcal{A}_i \} \right| \geq 2 \right\}$ \tcp*{Agreed issues}
$\mathcal{C} \leftarrow \left\{ a \mid a \in \bigcup_{i=1}^n \mathcal{A}_i \text{ and } \left| \{ i : a \in \mathcal{A}_i \} \right| < 2 \right\}$ \tcp*{Contested issues}

\textbf{Phase 3: Contested Issue Validation via Debate} \;
$\mathcal{K} \leftarrow \emptyset$ \tcp*{Initialize valid issues}
\For{each contested issue $c \in \mathcal{C}$}{
    \For{debate round $t = 1$ \textbf{to} $T_{\text{debate}}$}{
        \For{each reviewer $r_i \in \mathcal{R}$}{
            $\textsc{Argue}(r_i, c, D, z)$ \tcp*{Debate validity of issue $c$ with evidence from $D$ and $z$}
        }
    }
    $\text{vote}_c \leftarrow \textsc{MajorityVote}(\mathcal{R}, c)$ \tcp*{Vote on whether issue $c$ is valid}
    \If{$\text{vote}_c = \text{VALID}$}{
        $\mathcal{K} \leftarrow \mathcal{K} \cup \{c\}$ \tcp*{Keep valid contested issue}
    }
}

\textbf{Phase 4: Final Decision} \;
$\mathcal{I} \leftarrow \mathcal{M} \cup \mathcal{K}$ \tcp*{Combine major issues and kept contested issues into the final issue list}

\Return $\mathcal{I}$ \tcp*{Return issue list}
% \If{$\mathcal{I} = \emptyset$}{ 
%     $\mathit{dec} \leftarrow \text{ACCEPT}$ \tcp*{If no issues remain, accept the summary/quiz}
% }
% \Else{
%     $\mathit{dec} \leftarrow \text{REJECT}$ \tcp*{If issues remain, reject the summary/quiz}
% }

% \Return $(\mathit{dec}, \mathcal{I})$ \tcp*{Return the final decision and issue list}

\end{algorithm}

\subsection{Reviewer Collaboration}
\label{sec:method_reviewer}

The reviewer collaboration in \ours takes a systematic approach to quality assessment, where multiple reviewer agents work together to thoroughly evaluate generated summaries/quizzes and catch potential errors. As shown in \autoref{alg:reviewer}, this reviewing process also includes four phases:

\paragraph{Phase 1: Independent Reviewing} The reviewing process begins with each reviewer agent $r_i$ conducting an independent and comprehensive review of the generated summary/quiz $z$ against the original document $D$. During this phase, each reviewer meticulously examines the summary/quiz to identify various types of quality issues, including factual errors, omissions of important information, redundant content, and other potential problems. The reviewers produce individual annotation sets $\mathcal{A}_i$ that capture their unique perspectives and assessment criteria.

\paragraph{Phase 2: Issue Categorization} Once all reviewers have completed their independent reviews, the system categorizes the identified issues based on the level of agreement among reviewers. Issues that are flagged by at least two reviewers are classified as agreed issues $\mathcal{M}$, indicating a strong consensus that these problems genuinely exist. Conversely, issues identified by fewer than two reviewers are categorized as contested issues $\mathcal{C}$, suggesting disagreements that require further discussion.

\paragraph{Phase 3: Contested Issue Validation via Debate} For contested issues in $\mathcal{C}$, where initial reviewer agreement is lacking, \ours employs a structured debate mechanism to determine their validity. Each contested issue $c$ undergoes $T_{\texttt{debate}}$ rounds of debate, where all reviewer agents $r_i \in \mathcal{R}$ engage in evidence-based argumentation using the original document $D$ and summary or quiz $z$ as supporting materials. During each debate round, reviewers present their reasoning for or against the validity of issue $c$. 
After the debate rounds conclude, all reviewers participate in a majority vote to determine whether the contested issue should be considered valid. 

\paragraph{Phase 4: Final Decision} In the final phase, the reviewer collaboration reaches its final decision by consolidating all validated issues. The system combines the agreed issues $\mathcal{M}$ from Phase 2 with the valid contested issues $\mathcal{K}$ from Phase 3 to form the comprehensive final issue list $\mathcal{I} = \mathcal{M} \cup \mathcal{K}$. This issue list is returned to guide subsequent iterations of the generation process, ensuring that identified problems are systematically addressed in future revisions.

\section{Experiments}
\label{sec:experiments}

\subsection{Experimental Setup}
\label{sec:experimental_setup}

\paragraph{Evaluation}
We evaluate \ours on the long document summarization task using the \mensa \citep{saxena-keller-2024-select}, \booksum \citep{kryscinski-etal-2022-booksum}, and \govreport \citep{huang-etal-2021-efficient} benchmarks. Following standard protocols, we assess the generated summaries using ROUGE-1 (\rougeone), ROUGE-2 (\rougetwo), ROUGE-L (\rougel) \citep{lin-2004-rouge}, and BERTScore-$F_1$ (\bsf) \citep{DBLP:conf/iclr/ZhangKWWA20}.\footnote{BERTScore model: \texttt{bert-base-uncased}.} We also employ LLM-as-a-Judge evaluations with \gptfourone and \gptfive (detailed in \autoref{appsec:llm-as-judge}), and conduct human evaluations.  A case study is presented in \autoref{sec:appendix_case_study}.

\paragraph{Baselines}
We compare \ours against strong baselines from three categories: (1) \textbf{Supervised Fine-Tuning}: \textrank \citep{mihalcea2004textrank,jeong2025agent}, \longtfive \citep{guo2021longt5}, \unlim \citep{bertsch2023unlimiformer}, \sled \citep{ivgi2023efficient}, and \cached \citep{saxena2025end}. (2) \textbf{Prompting}: Proprietary LLMs including \gptfouro, \gptfourone, \gptfive, \gptothree, and open-source models such as \dpsk and \qwen. (3) \textbf{Multi-Agent Systems}: \hmsr \citep{jeong2025agent} and \cmllm \citep{fang2024multi}.

\paragraph{Ours}
In our experiments, we consider two configurations of our approach: \textbf{\ourssolo}, where each component employs a single agent for efficient and straightforward deployment, and \textbf{\oursensemble}, where each component leverages multiple agents in an ensemble manner to facilitate collaborative generation and review. 
By default, we use \gptfouro as the agent backbone for both \ourssolo and \oursensemble, deploying three agents in each component for \oursensemble unless otherwise specified. Moreover, the number of iterations $T_{\text{iter}}$ is set to three for all configurations and the number of debate rounds $T_{\text{debate}}$ is set to one. Prompts used in \ours are in \autoref{appsec:prompts}.

\subsection{Automatic Evaluation Results}
\label{sec:automatic_evaluation}

\begin{table}[t]
\centering
\small
\setlength{\tabcolsep}{3.5pt}
\caption{
Overall results given by different methods on \mensa, \booksum, and \govreport. The best results are highlighted in \textbf{bold}.}
\begin{tabular}{lcccccccccccc}
\toprule
& \multicolumn{4}{c}{\mensa}              & \multicolumn{4}{c}{\booksum}    & \multicolumn{4}{c}{\govreport}  \\ \cmidrule(rl){2-5} \cmidrule(rl){6-9} \cmidrule(rl){10-13}
& \rougeone       & \rougetwo       & \rougel   & \bsf & \rougeone   & \rougetwo   & \rougel   & \bsf & \rougeone   & \rougetwo   & \rougel   & \bsf \\ \midrule
\multicolumn{13}{c}{\cellcolor{gray!15}\textbf{Supervised Fine-Tuning}}                 \\
\textrank            & 34.37     & \phantom{0}4.60      & 12.84 & 48.10  &  -   &  -   &  -   &   - &  - & -   &  -  &   -  \\
\longtfive              & 20.77     & \phantom{0}2.26      & 10.03 & 45.01  & -  &  -  &  -  &        & -  &  -  &  -  &  -   \\
\unlim              &    -  &  -     & -   &  -   & 36.70 & \phantom{0}7.30  & 15.50 & 51.50  & 56.60 & \textbf{26.30} & 27.60 & \textbf{68.20}  \\
\sled                &  -     &   -   &  -  &  -  & 38.90 & \phantom{0}7.50  & 15.80 & 52.40  & \textbf{57.50} & \textbf{26.30} & 27.40 & 66.90  \\
\cached              &   -    &   -  &   -  &  -   & 42.80 & 10.50 & 18.80 & 54.40  & 57.00 & \textbf{26.30} & \textbf{28.19} & 67.30  \\
\multicolumn{13}{c}{\cellcolor{gray!15}\textbf{Prompting}}                 \\
\gptfouro               & 25.78     & \phantom{0}7.24      & 13.59 & 59.67  & 23.02 & \phantom{0}1.81  & 12.23 & 58.52  & 31.42 & 11.87 & 17.61 & 63.43  \\
\gptfourone              & 30.31     & \phantom{0}8.36      & 15.39 & 55.01  & 23.05 & \phantom{0}5.54  & 11.47 & 58.12  & 40.84 & 12.96 & 19.13 & 62.95  \\
\gptfive             & 37.38 & \phantom{0}9.14 & 17.11 & 60.44 & 23.98 & \phantom{0}5.69 & 12.38 & 58.55 & 41.52 & 12.52 & 19.23 & 62.55 \\ 
\gptothree                 & 32.84     & \phantom{0}8.54      & 17.09 & 59.27  & 22.00 & \phantom{0}5.24  & 11.51 & 58.44  & 38.28 & \phantom{0}9.93  & 17.47 & 61.19  \\
\dpsk           & 27.63     & \phantom{0}7.66      & 14.82 & 56.82  & 18.86 & \phantom{0}4.69  & \phantom{0}9.71  & 55.85  & 35.42 & \phantom{0}9.58  & 16.66 & 61.07  \\
\qwen               & 23.49     & \phantom{0}5.58      & 12.77 & 55.76  & 20.19 & \phantom{0}4.68  & 10.68 & 56.51  & 35.52 & 10.80 & 17.07 & 61.08  \\
\multicolumn{13}{c}{\cellcolor{gray!15}\textbf{Multi-Agent Systems}}               \\
\hmsr               & 34.26     & \phantom{0}9.74      & 13.46 & 60.22  & -   & -   &  -   &   -  &  -  & -  &  -  & -  \\
\cmllm               & -     & -      & - & -  & -   & -   &  -   &   -  &  47.90  & -  &  19.70  & -  \\ \midrule
\ourssolo          & 39.30     & \phantom{0}9.70      & 17.12 & 61.84  & 33.33 & \phantom{0}8.35  & 15.47 & 60.41  & 48.71 & 17.26 & 21.21 & 65.21  \\
\oursensemble     & \textbf{41.58}     & \textbf{10.96}     & \textbf{18.24} &\textbf{ 62.76 } & \textbf{44.62} & \textbf{11.14} & \textbf{20.38} & \textbf{61.49}  & 52.79 & 18.47 & 21.76 & 65.46  \\ \bottomrule
\end{tabular}
\label{tab:overall}
\end{table}

\paragraph{\oursensemble achieves strong performance with notable improvements on challenging datasets.}
\autoref{tab:overall} reports the automatic evaluation results of all methods on the \mensa, \booksum, and \govreport benchmarks. Our \oursensemble configuration demonstrates strong performance across all datasets, achieving the best results on \mensa and \booksum across all metrics. On \govreport, while some supervised fine-tuning baselines (\unlim, \sled, \cached) achieve competitive or superior performance on specific metrics due to their large-scale task-specific training, \oursensemble still outperforms all prompting baselines and shows substantial improvements over the \ourssolo variant. Notably, \oursensemble yields the most significant improvements on the challenging \booksum dataset, where it substantially outperforms all baselines across all metrics.
Furthermore, \ourssolo configuration also performs strongly, consistently surpassing prompting baselines.
These results confirm the advantages of our proposed multi-agent summarization framework, particularly the \oursensemble configuration, in handling diverse long documents.

\begin{figure}[t]
\centering          
\subfigure[\oursensemble vs. \gptfouro]{\label{fig:judge-act}\includegraphics[width=0.48\linewidth]{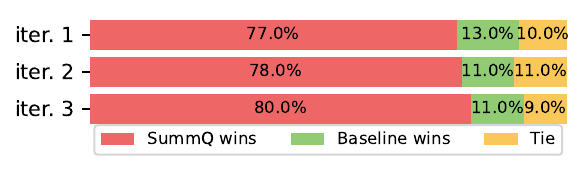}} 
\subfigure[\oursensemble vs. \gptothree]{\label{fig:judge-plan}\includegraphics[width=0.48\linewidth]{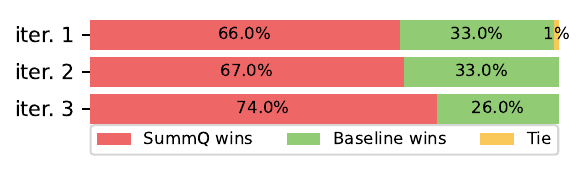}} 
\caption{The comparison between \oursensemble and baselines judged by \gptfive on \mensa during iteration, where there are three \gptfouro agents in each component of \oursensemble.}
\label{fig:llm-judge-evolve}
\end{figure}

\paragraph{LLM-as-a-Judge evaluation highlights the effectiveness of \oursensemble.} 
\autoref{fig:llm-judge-evolve} presents LLM-as-a-Judge results, comparing \oursensemble to strong baselines on the \mensa benchmark over multiple iterations. LLM judges (\gptfive) compare summary pairs and select the superior one, with each subfigure showing the winning rate of \oursensemble against different LLM agents (\gptfouro or \gptothree), judged by \gptfive. Across all settings, \oursensemble consistently outperforms baselines, achieving higher winning rates and demonstrating the effectiveness and generalizability of \oursensemble. 
The specific prompts and additional results judged by \gptfourone are in \autoref{appsec:llm-as-judge}.

\subsection{Human Evaluation Results}
\label{sec:human_evaluation}

\begin{figure}[t]
  \centering
  \begin{minipage}{0.43\linewidth}
    \centering
    \includegraphics[scale=0.55]{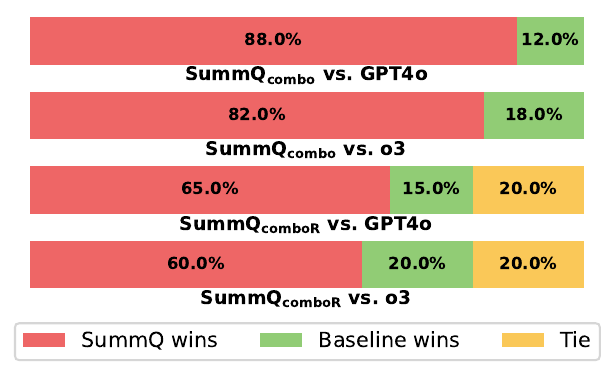}
    \caption{
    Human evaluation results comparing \gptfouro, \gptothree, \oursensemble, and \oursensemblerephrase. 
    }
    \label{fig:human_eval}
    \end{minipage}
    \hfill
    \begin{minipage}{0.56\linewidth}
        \captionsetup{type=table}
        \centering
\small
\setlength{\tabcolsep}{2pt}
\caption{Results on \mensa obtained by \ours, where one component contains multiple agents while other components contain only a single agent.}
\begin{tabular}{lcccccc}
\toprule
& \rougeone       & \rougetwo       & \rougel  &  \bsp & \bsr & \bsf \\ \midrule
\ourssolo &  39.30     & \phantom{0}9.70      & 17.12 & 62.19 &  61.55
 & 61.84  \\
\oursensemble            & \textbf{41.58} & \textbf{11.08}         & \textbf{18.24} & \textbf{63.28} & 62.28 & \textbf{62.76 }      \\ 
\multicolumn{7}{l}{\cellcolor{gray!15}\textbf{Only one component with 3 agents}}                 \\
Quiz Rev.     & 40.81          & 10.63          & 17.78          & 62.17          & 61.61          & 61.87          \\
Summary Rev.  & 41.20          & 10.80          & 17.93          & 62.29          & 61.72         & 61.99        \\
Quiz Gen.    & 40.40          & 10.80          & 17.95          & 62.54          & 61.72          & 62.11          \\
Summary Gen. & 40.72          & 10.96 & 18.07          & 62.70         & \textbf{62.39}          & 62.53 \\
\bottomrule  
\end{tabular}
\label{tab:isolation}

    \end{minipage}
\end{figure}

\paragraph{Setup}
We conduct a human evaluation using 20 randomly selected NLP papers published after June 2024, with five Ph.D. students as judges. Each judge compares two summaries considering \textit{Informativeness}, \textit{Coherence}, and \textit{Factuality}. The performance is measured by the winning rate. To address length bias, we include both \oursensemble and a rephrased version \oursensemblerephrase with shortened summaries. Details on the evaluation protocol and selected papers are in \autoref{appsec:human_eval}.

\paragraph{Results}
\autoref{fig:human_eval} presents the results of our human evaluation, comparing \oursensemble and \oursensemblerephrase against strong baselines, including \gptfouro and \gptothree. The results indicate that \oursensemble is preferred over \gptfouro and \gptothree with winning rates of 88\% and 82\%, respectively. Even after mitigating the potential length bias through rephrasing, \oursensemblerephrase still outperforms \gptfouro and \gptothree with winning rates of 65\% and 60\%, respectively. These findings underscore the effectiveness of our collaborative multi-agent framework in generating high-quality summaries that are favored by human judges, even when accounting for differences in summary length.

\section{Analysis}
\label{sec:analysis}

\paragraph{Multi-agent collaboration consistently excels for each component in \ours.} We analyze each component of \ours by replacing the single-agent setup with a multi-agent ensemble while keeping the other components as single-agent. As shown in \autoref{tab:isolation}, all components benefit from multi-agent collaboration, especially the \textit{Summary Generators} and \textit{Summary Reviewers}. These results highlight that collaboration and diverse perspectives significantly improve summary quality.

\begin{figure}[t]
  \centering
  \begin{minipage}{0.48\linewidth}
    \centering
    \captionsetup{type=table}
    
\centering
\small
\setlength{\tabcolsep}{2pt}
\caption{
Results with different number of iterations $T_{\text{iter}}$ on \mensa with the \oursensemble. All agents are \gptfouro.}
\begin{tabular}{ccccccc}
\toprule
\#iter. & \rougeone       & \rougetwo       & \rougel  &  \bsp & \bsr & \bsf \\ \midrule
1 & 38.14   & 10.44   & 17.85   & 62.77   & 61.50   & 62.60   \\
2 & 40.55   & 10.74   & 17.80   & 63.06   & 61.85   & 62.43   \\
3 & 41.58   & 11.08   & \textbf{18.24}   & \textbf{63.28}   & \textbf{62.28}   & \textbf{62.76 }  \\
4 & \textbf{41.62}   & \textbf{11.19} & 18.11   & 63.26   & 62.25   & 62.73   \\
5 & 41.53 & 11.01   & 18.21 & 62.90 &  62.23 & 62.55 \\ \bottomrule
\end{tabular}
\label{tab:num-rounds}

  \end{minipage}
  \hfill
  \begin{minipage}{0.48\linewidth}
    \centering
    \captionsetup{type=table}
    
\centering
\small
\setlength{\tabcolsep}{2pt}
\caption{
Results with different number of agents in each component on \mensa with the \oursensemble. All agents are \gptfouro.}
\begin{tabular}{ccccccc}
\toprule
\#agents& \rougeone       & \rougetwo       & \rougel  &  \bsp & \bsr & \bsf \\ \midrule
1 &  39.30     & \phantom{0}9.70      & 17.12 & 62.19 &  61.55
 & 61.84  \\
2 & 40.49          & 10.20          & 18.02          & 62.23         & 62.11         & 62.16          \\
3 & 41.58          & 11.08          & 18.24          & 63.28         & 62.28         & 62.76          \\
4 & 41.81          & 10.94          & 18.30          & 62.82          & 62.46         & 62.34         \\
5 & \textbf{42.52} & \textbf{11.49} & \textbf{18.56} & \textbf{63.53} & \textbf{62.99} & \textbf{62.96} \\ \bottomrule
\end{tabular}
\label{tab:num-agents}

  \end{minipage}
\end{figure}

\paragraph{More iterations of refinement does not always lead to better summaries.}
We analyze the effect of varying the number of iterations $T_{\text{iter}}$ in \ours (\autoref{alg:overall}) on summary quality for \mensa (\autoref{tab:num-rounds}). Performance generally improves from 1 to 3 iterations, with BERTScore-$F_1$ peaking at 62.76, but further iterations yield diminishing or negative returns. This suggests that too few iterations fail to fully leverage collaborative reasoning, while too many can introduce noise or over-refinement, indicating an optimal balance is needed.

\paragraph{More agents lead to better performance, but with diminishing returns and increased cost.}
As shown in \autoref{tab:num-agents}, increasing the number of agents in each component of \oursensemble generally improves performance, but gains become smaller as more agents are added. For example, ROUGE-L rises from 17.12 (one agent) to 18.02 (two agents), but further increases yield only minor improvements. This highlights a trade-off between summarization quality and managing computational cost, as adding agents increases costs without proportional benefits.

\begin{table}[t]
\centering
\small
\setlength{\tabcolsep}{3.2pt}
\caption{
Results given by \oursensemble with different LLMs as agent backbone on \mensa.}
\begin{tabular}{lcccccclcccccc}
\toprule
\multicolumn{7}{c}{Proprietary Models}                                                     & \multicolumn{7}{c}{Open-source Models}                    \\ \cmidrule(rl){1-7} \cmidrule(rl){8-14}
         & \rougeone       & \rougetwo            & \rougel       & \bsp     & \bsr     & \bsf         &           & \rougeone   & \rougetwo  & \rougel   & \bsp & \bsr & \bsf \\ \midrule
\multicolumn{7}{l}{\cellcolor{gray!15}\gptfouromini}                                                            & \multicolumn{7}{l}{\cellcolor{gray!15}\dpsk}                           \\
baseline & 26.61     & \phantom{0}6.56           & 13.77     & 57.66     & 55.87     & 58.15          & baseline  & 27.63 & \phantom{0}7.66 & 14.82 & 54.98 & 58.86 & 56.82   \\
\ours     & 35.18     & \phantom{0}7.97           & 15.67    & 58.02 & 58.64 & 58.31       & \ours     & 30.71 & \phantom{0}7.77 & 15.04 & 61.99 & 58.77 & 60.30  \\
\multicolumn{7}{l}{\cellcolor{gray!15}\gptfourone}                                                                & \multicolumn{7}{l}{\cellcolor{gray!15}\qwen}  \\
baseline & 30.31     & \phantom{0}8.36           & 15.39     & 56.00    & 54.12     & 55.01          & baseline   & 23.49 & \phantom{0}5.58 & 12.77 & 55.86 & 55.99 & 55.76  \\
\ours     & 49.17     & 12.25 & 18.98     & 59.58     & 62.42     & 60.95 & \ours     & 26.50 & \phantom{0}5.80 & 13.14 & 57.13 & 59.09 & 58.02       \\
\multicolumn{7}{l}{\cellcolor{gray!15}O3}                                                                     & \multicolumn{7}{l}{\cellcolor{gray!15}\model{DeepSeek-R1-Distill-Qwen-32B}}          \\
baseline &32.84  & 	\phantom{0}8.54 & 17.09 & 57.38 & 59.81 & 59.27     &  baseline   & 26.66 & \phantom{0}6.35 & 13.77 & 58.83 & 55.57 & 57.07     \\
\ours     & 46.69     & 10.37          & 19.09     & 61.90     & 61.30     & 60.82          & \ours    & 31.07 & \phantom{0}6.99 & 14.70 & 58.80 & 57.15 & 57.83     \\ \bottomrule
\end{tabular}
\label{tab:models}
\end{table}

\paragraph{\ours consistently achieves superior performance with diverse LLM agent backbones.}
\autoref{tab:models} shows that \oursensemble outperforms all baselines across various proprietary and open-source LLM backbones. Summarization quality strongly depends on the agent backbone: advanced models like \gptfourone and \gptothree perform better than their weaker counterparts. This highlights the importance of choosing robust LLMs to maximize multi-agent summarization performance. We also explore the impact of combining different LLMs within \ours in \autoref{appsec:combination}.

\begin{wraptable}[10]{t}{7cm}
\centering
\small
\setlength{\tabcolsep}{1pt}
\caption{
Results with and without quiz on \mensa.}
\begin{tabular}{lcccccc}
\toprule
& \rougeone       & \rougetwo       & \rougel  &  \bsp & \bsr & \bsf \\ \midrule
\ourssolo &  39.30     & \phantom{0}9.70      & 17.12 & 62.19 &  61.55
 & 61.84  \\
├ w/o quiz  & 34.30  & \phantom{0}8.99  & 15.85  & 59.71   & 51.80  & 55.44  \\ \midrule
\oursensemble            & 41.58 & 11.08         & 18.24 & 63.28 & 62.28 & 62.76       \\ 
├ w/o quiz  & 39.49  &  \phantom{0}9.33 & 17.13  &  61.10  & 60.39  & 60.59  \\ 
\bottomrule  
\end{tabular}
\label{tab:wo-quiz}
\end{wraptable}
\paragraph{Quizzing mechanism effectively improves the summarization quality.}
To evaluate the contribution of the quizzing mechanism in \ours, we ablate the quizzing mechanism in \ours by removing generators and reviewers for quizzing and the examinee. As shown in \autoref{tab:wo-quiz}, this leads to consistent performance drops across all metrics on \mensa. For \ourssolo, \rougeone drops by 5.00 and \bsf by 6.40; for \oursensemble, \rougeone and \bsf decrease by 2.09 and 2.17, respectively. These results show that the quizzing mechanism is crucial for enhancing summarization quality by comprehensively challenging the generated summaries.

\begin{wrapfigure}[9]{r}{5cm}
  \centering
    \includegraphics[scale=0.55]{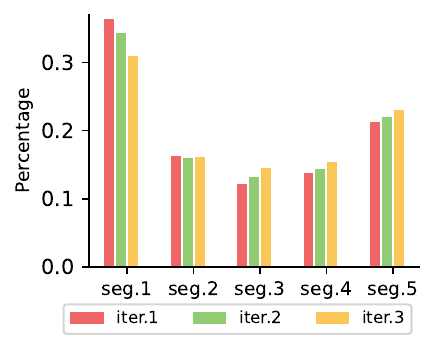}
    \caption{
    Quiz question distribution on \mensa with the \oursensemble. 
    }
    \label{fig:quiz-distribution}
\end{wrapfigure}

\paragraph{Quiz coverage gets more balanced across document segments as the iteration proceeds.}
We divide each document into five equal segments and use \gptfourone to map quiz questions to segments. As shown in \autoref{fig:quiz-distribution}, quiz questions initially focus on the beginning and end, which aligns well with the findings of \citet{liu-etal-2024-lost}, but coverage becomes more balanced by iteration 3, with increased attention to middle segments. This shift indicates a more holistic document understanding, which is crucial for generating summaries that accurately reflect the entire content.
\section{Conclusion}

In this work, we introduce \ours, a novel adversarial multi-agent framework that addresses critical challenges in long document summarization through collaborative intelligence between specialized summarization and quizzing agents. Our approach creates a natural adversarial dynamic where quiz generation serves as a continuous quality check, ensuring comprehensive coverage, factual accuracy, and verifiability of summaries through iterative refinement. Extensive experiments on three benchmarks demonstrate that \ours achieves superior performance across multiple evaluation metrics including ROUGE, BERTScore, LLM-as-a-judge, and human assessments. Our comprehensive analyses reveal the effectiveness of multi-agent collaboration, the impact of the quizzing mechanism on summary quality, and the influence of various agent configurations.

\section*{Ethics Statement}
This work introduces \ours, an adversarial multi-agent framework for long document summarization using large language models (LLMs). All experiments were conducted using publicly available datasets and LLMs, strictly adhering to their respective licenses and usage policies. No human subjects were involved in the development or evaluation of the system, except for the human evaluation, which was performed by consenting Ph.D. students with relevant expertise. We acknowledge that LLMs and datasets may contain inherent biases, which could affect the generated summaries and quiz questions. We encourage responsible use of our framework, with attention to fairness, transparency, and accountability in downstream applications.

\section*{Reproducibility Statement}
We are committed to reproducibility in this work. Detailed descriptions of the \ours framework, including algorithms, agent configurations, and collaboration mechanisms, are provided in \autoref{sec:method}. Experimental setups, including model backbones, datasets, evaluation metrics, and baseline comparisons, are thoroughly described in \autoref{sec:experiments}. All datasets used are standard benchmarks, and references are included for accessibility. Prompts and implementation details are provided in Appendix. To further support reproducibility, we will release our code and experiment scripts upon publication, enabling other researchers to replicate and extend our results.

\section*{The Use of Large Language Models (LLMs)}

In preparing this work, we utilize large language models (LLMs) as general-purpose tools to assist with writing polish and grammar correction. The LLMs are not involved in research ideation, experimental design, or substantive content generation. Their role is limited to improving the clarity and readability of the text, ensuring grammatical accuracy, and refining the presentation of our findings. All scientific contributions, analyses, and conclusions are solely the work of the authors.

\section*{Acknowledgement}

This project has received funding from UK Research and Innovation (UKRI) under the UK government’s Horizon Europe funding guarantee [grant numbers 10039436].

The computations described in this research were performed using the Baskerville Tier 2 HPC service (https://www.baskerville.ac.uk/). Baskerville was funded by the EPSRC and UKRI through the World Class Labs scheme (EP/T022221/1) and the Digital Research Infrastructure programme (EP/W032244/1) and is operated by Advanced Research Computing at the University of Birmingham.
\bibliography{iclr2026_conference}

@inproceedings{fabbri-etal-2019-multi,
    title = "Multi-News: A Large-Scale Multi-Document Summarization Dataset and Abstractive Hierarchical Model",
    author = "Fabbri, Alexander  and
      Li, Irene  and
      She, Tianwei  and
      Li, Suyi  and
      Radev, Dragomir",
    editor = "Korhonen, Anna  and
      Traum, David  and
      M{\`a}rquez, Llu{\'i}s",
    booktitle = "Proceedings of the 57th Annual Meeting of the Association for Computational Linguistics",
    month = jul,
    year = "2019",
    address = "Florence, Italy",
    publisher = "Association for Computational Linguistics",
    url = "https://aclanthology.org/P19-1102/",
    doi = "10.18653/v1/P19-1102",
    pages = "1074--1084"
}

@inproceedings{see-etal-2017-get,
    title = "Get To The Point: Summarization with Pointer-Generator Networks",
    author = "See, Abigail  and
      Liu, Peter J.  and
      Manning, Christopher D.",
    editor = "Barzilay, Regina  and
      Kan, Min-Yen",
    booktitle = "Proceedings of the 55th Annual Meeting of the Association for Computational Linguistics (Volume 1: Long Papers)",
    month = jul,
    year = "2017",
    address = "Vancouver, Canada",
    publisher = "Association for Computational Linguistics",
    url = "https://aclanthology.org/P17-1099/",
    doi = "10.18653/v1/P17-1099",
    pages = "1073--1083"
}

@article{DBLP:journals/corr/abs-2405-11804,
  author       = {Minghao Wu and
                  Yulin Yuan and
                  Gholamreza Haffari and
                  Longyue Wang},
  title        = {(Perhaps) Beyond Human Translation: Harnessing Multi-Agent Collaboration
                  for Translating Ultra-Long Literary Texts},
  journal      = {CoRR},
  volume       = {abs/2405.11804},
  year         = {2024},
  url          = {https://doi.org/10.48550/arXiv.2405.11804},
  doi          = {10.48550/ARXIV.2405.11804},
  eprinttype    = {arXiv},
  eprint       = {2405.11804},
  bibsource    = {dblp computer science bibliography, https://dblp.org}
}

@inproceedings{kim-kim-2025-nexussum,
    title = "{N}exus{S}um: Hierarchical {LLM} Agents for Long-Form Narrative Summarization",
    author = "Kim, Hyuntak  and
      Kim, Byung-Hak",
    editor = "Che, Wanxiang  and
      Nabende, Joyce  and
      Shutova, Ekaterina  and
      Pilehvar, Mohammad Taher",
    booktitle = "Proceedings of the 63rd Annual Meeting of the Association for Computational Linguistics (Volume 1: Long Papers)",
    month = jul,
    year = "2025",
    address = "Vienna, Austria",
    publisher = "Association for Computational Linguistics",
    url = "https://aclanthology.org/2025.acl-long.500/",
    doi = "10.18653/v1/2025.acl-long.500",
    pages = "10120--10157",
    ISBN = "979-8-89176-251-0"
}

@inproceedings{xia-etal-2024-hallucination,
    title = "Hallucination Diversity-Aware Active Learning for Text Summarization",
    author = "Xia, Yu  and
      Liu, Xu  and
      Yu, Tong  and
      Kim, Sungchul  and
      Rossi, Ryan  and
      Rao, Anup  and
      Mai, Tung  and
      Li, Shuai",
    editor = "Duh, Kevin  and
      Gomez, Helena  and
      Bethard, Steven",
    booktitle = "Proceedings of the 2024 Conference of the North American Chapter of the Association for Computational Linguistics: Human Language Technologies (Volume 1: Long Papers)",
    month = jun,
    year = "2024",
    address = "Mexico City, Mexico",
    publisher = "Association for Computational Linguistics",
    url = "https://aclanthology.org/2024.naacl-long.479/",
    doi = "10.18653/v1/2024.naacl-long.479",
    pages = "8665--8677"
}

@article{chrysostomou-etal-2024-investigating,
    title = "Investigating Hallucinations in Pruned Large Language Models for Abstractive Summarization",
    author = "Chrysostomou, George  and
      Zhao, Zhixue  and
      Williams, Miles  and
      Aletras, Nikolaos",
    journal = "Transactions of the Association for Computational Linguistics",
    volume = "12",
    year = "2024",
    address = "Cambridge, MA",
    publisher = "MIT Press",
    url = "https://aclanthology.org/2024.tacl-1.64/",
    doi = "10.1162/tacl_a_00695",
    pages = "1163--1181"
}

@inproceedings{tang-etal-2024-tofueval,
    title = "{T}ofu{E}val: Evaluating Hallucinations of {LLM}s on Topic-Focused Dialogue Summarization",
    author = "Tang, Liyan  and
      Shalyminov, Igor  and
      Wong, Amy  and
      Burnsky, Jon  and
      Vincent, Jake  and
      Yang, Yu{'}an  and
      Singh, Siffi  and
      Feng, Song  and
      Song, Hwanjun  and
      Su, Hang  and
      Sun, Lijia  and
      Zhang, Yi  and
      Mansour, Saab  and
      McKeown, Kathleen",
    editor = "Duh, Kevin  and
      Gomez, Helena  and
      Bethard, Steven",
    booktitle = "Proceedings of the 2024 Conference of the North American Chapter of the Association for Computational Linguistics: Human Language Technologies (Volume 1: Long Papers)",
    month = jun,
    year = "2024",
    address = "Mexico City, Mexico",
    publisher = "Association for Computational Linguistics",
    url = "https://aclanthology.org/2024.naacl-long.251/",
    doi = "10.18653/v1/2024.naacl-long.251",
    pages = "4455--4480"
}

@article{DBLP:journals/csur/KohJLP23,
  author       = {Huan Yee Koh and
                  Jiaxin Ju and
                  Ming Liu and
                  Shirui Pan},
  title        = {An Empirical Survey on Long Document Summarization: Datasets, Models,
                  and Metrics},
  journal      = {{ACM} Comput. Surv.},
  volume       = {55},
  number       = {8},
  pages        = {154:1--154:35},
  year         = {2023},
  url          = {https://doi.org/10.1145/3545176},
  doi          = {10.1145/3545176},
  bibsource    = {dblp computer science bibliography, https://dblp.org}
}

@article{DBLP:journals/air/GambhirG17,
  author       = {Mahak Gambhir and
                  Vishal Gupta},
  title        = {Recent automatic text summarization techniques: a survey},
  journal      = {Artif. Intell. Rev.},
  volume       = {47},
  number       = {1},
  pages        = {1--66},
  year         = {2017},
  url          = {https://doi.org/10.1007/s10462-016-9475-9},
  doi          = {10.1007/S10462-016-9475-9},
  bibsource    = {dblp computer science bibliography, https://dblp.org}
}

@inproceedings{laban-etal-2023-summedits,
    title = "{S}umm{E}dits: Measuring {LLM} Ability at Factual Reasoning Through The Lens of Summarization",
    author = "Laban, Philippe  and
      Kryscinski, Wojciech  and
      Agarwal, Divyansh  and
      Fabbri, Alexander  and
      Xiong, Caiming  and
      Joty, Shafiq  and
      Wu, Chien-Sheng",
    editor = "Bouamor, Houda  and
      Pino, Juan  and
      Bali, Kalika",
    booktitle = "Proceedings of the 2023 Conference on Empirical Methods in Natural Language Processing",
    month = dec,
    year = "2023",
    address = "Singapore",
    publisher = "Association for Computational Linguistics",
    url = "https://aclanthology.org/2023.emnlp-main.600/",
    doi = "10.18653/v1/2023.emnlp-main.600",
    pages = "9662--9676"
}

@article{liu-etal-2024-lost,
    title = "Lost in the Middle: How Language Models Use Long Contexts",
    author = "Liu, Nelson F.  and
      Lin, Kevin  and
      Hewitt, John  and
      Paranjape, Ashwin  and
      Bevilacqua, Michele  and
      Petroni, Fabio  and
      Liang, Percy",
    journal = "Transactions of the Association for Computational Linguistics",
    volume = "12",
    year = "2024",
    address = "Cambridge, MA",
    publisher = "MIT Press",
    url = "https://aclanthology.org/2024.tacl-1.9/",
    doi = "10.1162/tacl_a_00638",
    pages = "157--173"
}

@inproceedings{kryscinski-etal-2022-booksum,
    title = "{BOOKSUM}: A Collection of Datasets for Long-form Narrative Summarization",
    author = "Kryscinski, Wojciech  and
      Rajani, Nazneen  and
      Agarwal, Divyansh  and
      Xiong, Caiming  and
      Radev, Dragomir",
    editor = "Goldberg, Yoav  and
      Kozareva, Zornitsa  and
      Zhang, Yue",
    booktitle = "Findings of the Association for Computational Linguistics: EMNLP 2022",
    month = dec,
    year = "2022",
    address = "Abu Dhabi, United Arab Emirates",
    publisher = "Association for Computational Linguistics",
    url = "https://aclanthology.org/2022.findings-emnlp.488/",
    doi = "10.18653/v1/2022.findings-emnlp.488",
    pages = "6536--6558",
}

@inproceedings{saxena-keller-2024-select,
    title = "Select and Summarize: Scene Saliency for Movie Script Summarization",
    author = "Saxena, Rohit  and
      Keller, Frank",
    editor = "Duh, Kevin  and
      Gomez, Helena  and
      Bethard, Steven",
    booktitle = "Findings of the Association for Computational Linguistics: NAACL 2024",
    month = jun,
    year = "2024",
    address = "Mexico City, Mexico",
    publisher = "Association for Computational Linguistics",
    url = "https://aclanthology.org/2024.findings-naacl.218/",
    doi = "10.18653/v1/2024.findings-naacl.218",
    pages = "3439--3455",
}

@inproceedings{huang-etal-2021-efficient,
    title = "Efficient Attentions for Long Document Summarization",
    author = "Huang, Luyang  and
      Cao, Shuyang  and
      Parulian, Nikolaus  and
      Ji, Heng  and
      Wang, Lu",
    editor = "Toutanova, Kristina  and
      Rumshisky, Anna  and
      Zettlemoyer, Luke  and
      Hakkani-Tur, Dilek  and
      Beltagy, Iz  and
      Bethard, Steven  and
      Cotterell, Ryan  and
      Chakraborty, Tanmoy  and
      Zhou, Yichao",
    booktitle = "Proceedings of the 2021 Conference of the North American Chapter of the Association for Computational Linguistics: Human Language Technologies",
    month = jun,
    year = "2021",
    address = "Online",
    publisher = "Association for Computational Linguistics",
    url = "https://aclanthology.org/2021.naacl-main.112/",
    doi = "10.18653/v1/2021.naacl-main.112",
    pages = "1419--1436",
}

@inproceedings{lin-2004-rouge,
    title = "{ROUGE}: A Package for Automatic Evaluation of Summaries",
    author = "Lin, Chin-Yew",
    booktitle = "Text Summarization Branches Out",
    month = jul,
    year = "2004",
    address = "Barcelona, Spain",
    publisher = "Association for Computational Linguistics",
    url = "https://aclanthology.org/W04-1013/",
    pages = "74--81"
}

@inproceedings{DBLP:conf/iclr/ZhangKWWA20,
  author       = {Tianyi Zhang and
                  Varsha Kishore and
                  Felix Wu and
                  Kilian Q. Weinberger and
                  Yoav Artzi},
  title        = {BERTScore: Evaluating Text Generation with {BERT}},
  booktitle    = {8th International Conference on Learning Representations, {ICLR} 2020,
                  Addis Ababa, Ethiopia, April 26-30, 2020},
  publisher    = {OpenReview.net},
  year         = {2020},
  url          = {https://openreview.net/forum?id=SkeHuCVFDr},
  bibsource    = {dblp computer science bibliography, https://dblp.org}
}

@article{bertsch2023unlimiformer,
  title={Unlimiformer: Long-range transformers with unlimited length input},
  author={Bertsch, Amanda and Alon, Uri and Neubig, Graham and Gormley, Matthew},
  journal={Advances in Neural Information Processing Systems},
  volume={36},
  pages={35522--35543},
  year={2023}
}

@article{saxena2025end,
  title={End-to-End Long Document Summarization using Gradient Caching},
  author={Saxena, Rohit and Tang, Hao and Keller, Frank},
  journal={arXiv preprint arXiv:2501.01805},
  year={2025}
}

@article{ivgi2023efficient,
  title={Efficient long-text understanding with short-text models},
  author={Ivgi, Maor and Shaham, Uri and Berant, Jonathan},
  journal={Transactions of the Association for Computational Linguistics},
  volume={11},
  pages={284--299},
  year={2023},
  publisher={MIT Press One Broadway, 12th Floor, Cambridge, Massachusetts 02142, USA~…}
}

@inproceedings{mihalcea2004textrank,
  title={Textrank: Bringing order into text},
  author={Mihalcea, Rada and Tarau, Paul},
  booktitle={Proceedings of the 2004 conference on empirical methods in natural language processing},
  pages={404--411},
  year={2004}
}

@article{guo2021longt5,
  title={LongT5: Efficient text-to-text transformer for long sequences},
  author={Guo, Mandy and Ainslie, Joshua and Uthus, David and Ontanon, Santiago and Ni, Jianmo and Sung, Yun-Hsuan and Yang, Yinfei},
  journal={arXiv preprint arXiv:2112.07916},
  year={2021}
}

@article{jeong2025agent,
  title={Agent-as-judge for factual summarization of long narratives},
  author={Jeong, Yeonseok and Kim, Minsoo and Hwang, Seung-won and Kim, Byung-Hak},
  journal={arXiv preprint arXiv:2501.09993},
  year={2025}
}

@article{fang2024multi,
  title={Multi-llm text summarization},
  author={Fang, Jiangnan and Liu, Cheng-Tse and Kim, Jieun and Bhedaru, Yash and Liu, Ethan and Singh, Nikhil and Lipka, Nedim and Mathur, Puneet and Ahmed, Nesreen K and Dernoncourt, Franck and others},
  journal={arXiv preprint arXiv:2412.15487},
  year={2024}
}

@inproceedings{DBLP:conf/ijcai/GuoCWCPCW024,
  author       = {Taicheng Guo and
                  Xiuying Chen and
                  Yaqi Wang and
                  Ruidi Chang and
                  Shichao Pei and
                  Nitesh V. Chawla and
                  Olaf Wiest and
                  Xiangliang Zhang},
  title        = {Large Language Model Based Multi-agents: {A} Survey of Progress and
                  Challenges},
  booktitle    = {Proceedings of the Thirty-Third International Joint Conference on
                  Artificial Intelligence, {IJCAI} 2024, Jeju, South Korea, August 3-9,
                  2024},
  pages        = {8048--8057},
  publisher    = {ijcai.org},
  year         = {2024},
  url          = {https://www.ijcai.org/proceedings/2024/890},
  bibsource    = {dblp computer science bibliography, https://dblp.org}
}

@inproceedings{DBLP:conf/icml/Du00TM24,
  author       = {Yilun Du and
                  Shuang Li and
                  Antonio Torralba and
                  Joshua B. Tenenbaum and
                  Igor Mordatch},
  title        = {Improving Factuality and Reasoning in Language Models through Multiagent
                  Debate},
  booktitle    = {Forty-first International Conference on Machine Learning, {ICML} 2024,
                  Vienna, Austria, July 21-27, 2024},
  publisher    = {OpenReview.net},
  year         = {2024},
  url          = {https://openreview.net/forum?id=zj7YuTE4t8},
  bibsource    = {dblp computer science bibliography, https://dblp.org}
}

@inproceedings{DBLP:conf/emnlp/XiongD00023,
  author       = {Kai Xiong and
                  Xiao Ding and
                  Yixin Cao and
                  Ting Liu and
                  Bing Qin},
  editor       = {Houda Bouamor and
                  Juan Pino and
                  Kalika Bali},
  title        = {Examining Inter-Consistency of Large Language Models Collaboration:
                  An In-depth Analysis via Debate},
  booktitle    = {Findings of the Association for Computational Linguistics: {EMNLP}
                  2023, Singapore, December 6-10, 2023},
  pages        = {7572--7590},
  publisher    = {Association for Computational Linguistics},
  year         = {2023},
  url          = {https://doi.org/10.18653/v1/2023.findings-emnlp.508},
  doi          = {10.18653/V1/2023.FINDINGS-EMNLP.508},
  bibsource    = {dblp computer science bibliography, https://dblp.org}
}

@article{DBLP:journals/corr/abs-2310-20151,
  author       = {Huaben Chen and
                  Wenkang Ji and
                  Lufeng Xu and
                  Shiyu Zhao},
  title        = {Multi-Agent Consensus Seeking via Large Language Models},
  journal      = {CoRR},
  volume       = {abs/2310.20151},
  year         = {2023},
  url          = {https://doi.org/10.48550/arXiv.2310.20151},
  doi          = {10.48550/ARXIV.2310.20151},
  eprinttype    = {arXiv},
  eprint       = {2310.20151},
  bibsource    = {dblp computer science bibliography, https://dblp.org}
}

@inproceedings{DBLP:conf/acl/TangZ0L0ZCG24,
  author       = {Xiangru Tang and
                  Anni Zou and
                  Zhuosheng Zhang and
                  Ziming Li and
                  Yilun Zhao and
                  Xingyao Zhang and
                  Arman Cohan and
                  Mark Gerstein},
  editor       = {Lun{-}Wei Ku and
                  Andre Martins and
                  Vivek Srikumar},
  title        = {MedAgents: Large Language Models as Collaborators for Zero-shot Medical
                  Reasoning},
  booktitle    = {Findings of the Association for Computational Linguistics, {ACL} 2024,
                  Bangkok, Thailand and virtual meeting, August 11-16, 2024},
  pages        = {599--621},
  publisher    = {Association for Computational Linguistics},
  year         = {2024},
  url          = {https://doi.org/10.18653/v1/2024.findings-acl.33},
  doi          = {10.18653/V1/2024.FINDINGS-ACL.33},
  bibsource    = {dblp computer science bibliography, https://dblp.org}
}

@inproceedings{DBLP:conf/acl/ChenSB24,
  author       = {Justin Chih{-}Yao Chen and
                  Swarnadeep Saha and
                  Mohit Bansal},
  editor       = {Lun{-}Wei Ku and
                  Andre Martins and
                  Vivek Srikumar},
  title        = {ReConcile: Round-Table Conference Improves Reasoning via Consensus
                  among Diverse LLMs},
  booktitle    = {Proceedings of the 62nd Annual Meeting of the Association for Computational
                  Linguistics (Volume 1: Long Papers), {ACL} 2024, Bangkok, Thailand,
                  August 11-16, 2024},
  pages        = {7066--7085},
  publisher    = {Association for Computational Linguistics},
  year         = {2024},
  url          = {https://doi.org/10.18653/v1/2024.acl-long.381},
  doi          = {10.18653/V1/2024.ACL-LONG.381},
  bibsource    = {dblp computer science bibliography, https://dblp.org}
}

@inproceedings{DBLP:conf/naacl/SahaLCBWL24,
  author       = {Swarnadeep Saha and
                  Omer Levy and
                  Asli Celikyilmaz and
                  Mohit Bansal and
                  Jason Weston and
                  Xian Li},
  editor       = {Kevin Duh and
                  Helena G{\'{o}}mez{-}Adorno and
                  Steven Bethard},
  title        = {Branch-Solve-Merge Improves Large Language Model Evaluation and Generation},
  booktitle    = {Proceedings of the 2024 Conference of the North American Chapter of
                  the Association for Computational Linguistics: Human Language Technologies
                  (Volume 1: Long Papers), {NAACL} 2024, Mexico City, Mexico, June 16-21,
                  2024},
  pages        = {8352--8370},
  publisher    = {Association for Computational Linguistics},
  year         = {2024},
  url          = {https://doi.org/10.18653/v1/2024.naacl-long.462},
  doi          = {10.18653/V1/2024.NAACL-LONG.462},
  bibsource    = {dblp computer science bibliography, https://dblp.org}
}

@article{DBLP:journals/corr/abs-2311-08152,
  author       = {Zhenran Xu and
                  Senbao Shi and
                  Baotian Hu and
                  Jindi Yu and
                  Dongfang Li and
                  Min Zhang and
                  Yuxiang Wu},
  title        = {Towards Reasoning in Large Language Models via Multi-Agent Peer Review
                  Collaboration},
  journal      = {CoRR},
  volume       = {abs/2311.08152},
  year         = {2023},
  url          = {https://doi.org/10.48550/arXiv.2311.08152},
  doi          = {10.48550/ARXIV.2311.08152},
  eprinttype    = {arXiv},
  eprint       = {2311.08152},
  bibsource    = {dblp computer science bibliography, https://dblp.org}
}

@article{wang2025odysseybench,
  title={OdysseyBench: Evaluating LLM Agents on Long-Horizon Complex Office Application Workflows},
  author={Wang, Weixuan and Han, Dongge and Diaz, Daniel Madrigal and Xu, Jin and R{\"u}hle, Victor and Rajmohan, Saravan},
  journal={arXiv preprint arXiv:2508.09124},
  year={2025}
}

@article{DBLP:journals/corr/abs-2312-13010,
  author       = {Dong Huang and
                  Qingwen Bu and
                  Jie M. Zhang and
                  Michael Luck and
                  Heming Cui},
  title        = {AgentCoder: Multi-Agent-based Code Generation with Iterative Testing
                  and Optimisation},
  journal      = {CoRR},
  volume       = {abs/2312.13010},
  year         = {2023},
  url          = {https://doi.org/10.48550/arXiv.2312.13010},
  doi          = {10.48550/ARXIV.2312.13010},
  eprinttype    = {arXiv},
  eprint       = {2312.13010},
  bibsource    = {dblp computer science bibliography, https://dblp.org}
}

@inproceedings{DBLP:conf/coling/WangR0LBCYZYSL25,
  author       = {Bing Wang and
                  Changyu Ren and
                  Jian Yang and
                  Xinnian Liang and
                  Jiaqi Bai and
                  Linzheng Chai and
                  Zhao Yan and
                  Qian{-}Wen Zhang and
                  Di Yin and
                  Xing Sun and
                  Zhoujun Li},
  editor       = {Owen Rambow and
                  Leo Wanner and
                  Marianna Apidianaki and
                  Hend Al{-}Khalifa and
                  Barbara Di Eugenio and
                  Steven Schockaert},
  title        = {{MAC-SQL:} {A} Multi-Agent Collaborative Framework for Text-to-SQL},
  booktitle    = {Proceedings of the 31st International Conference on Computational
                  Linguistics, {COLING} 2025, Abu Dhabi, UAE, January 19-24, 2025},
  pages        = {540--557},
  publisher    = {Association for Computational Linguistics},
  year         = {2025},
  url          = {https://aclanthology.org/2025.coling-main.36/},
  bibsource    = {dblp computer science bibliography, https://dblp.org}
}

@article{wang2024survey,
  title={A survey on large language model based autonomous agents},
  author={Wang, Lei and Ma, Chen and Feng, Xueyang and Zhang, Zeyu and Yang, Hao and Zhang, Jingsen and Chen, Zhiyuan and Tang, Jiakai and Chen, Xu and Lin, Yankai and others},
  journal={Frontiers of Computer Science},
  volume={18},
  number={6},
  pages={186345},
  year={2024},
  publisher={Springer}
}

@article{xi2025rise,
  title={The rise and potential of large language model based agents: A survey},
  author={Xi, Zhiheng and Chen, Wenxiang and Guo, Xin and He, Wei and Ding, Yiwen and Hong, Boyang and Zhang, Ming and Wang, Junzhe and Jin, Senjie and Zhou, Enyu and others},
  journal={Science China Information Sciences},
  volume={68},
  number={2},
  pages={121101},
  year={2025},
  publisher={Springer}
}

@article{DBLP:journals/corr/abs-2004-05150,
  author       = {Iz Beltagy and
                  Matthew E. Peters and
                  Arman Cohan},
  title        = {Longformer: The Long-Document Transformer},
  journal      = {CoRR},
  volume       = {abs/2004.05150},
  year         = {2020},
  url          = {https://arxiv.org/abs/2004.05150},
  eprinttype    = {arXiv},
  eprint       = {2004.05150},
  bibsource    = {dblp computer science bibliography, https://dblp.org}
}

@inproceedings{DBLP:conf/emnlp/0006ZWX0Y21,
  author       = {Ye Liu and
                  Jian{-}Guo Zhang and
                  Yao Wan and
                  Congying Xia and
                  Lifang He and
                  Philip S. Yu},
  editor       = {Marie{-}Francine Moens and
                  Xuanjing Huang and
                  Lucia Specia and
                  Scott Wen{-}tau Yih},
  title        = {{HETFORMER:} Heterogeneous Transformer with Sparse Attention for Long-Text
                  Extractive Summarization},
  booktitle    = {Proceedings of the 2021 Conference on Empirical Methods in Natural
                  Language Processing, {EMNLP} 2021, Virtual Event / Punta Cana, Dominican
                  Republic, 7-11 November, 2021},
  pages        = {146--154},
  publisher    = {Association for Computational Linguistics},
  year         = {2021},
  url          = {https://doi.org/10.18653/v1/2021.emnlp-main.13},
  doi          = {10.18653/V1/2021.EMNLP-MAIN.13},
  bibsource    = {dblp computer science bibliography, https://dblp.org}
}

@article{liu2023lost,
  title={Lost in the middle: How language models use long contexts},
  author={Liu, Nelson F and Lin, Kevin and Hewitt, John and Paranjape, Ashwin and Bevilacqua, Michele and Petroni, Fabio and Liang, Percy},
  journal={arXiv preprint arXiv:2307.03172},
  year={2023}
}

@article{DBLP:journals/corr/abs-2306-15595,
  author       = {Shouyuan Chen and
                  Sherman Wong and
                  Liangjian Chen and
                  Yuandong Tian},
  title        = {Extending Context Window of Large Language Models via Positional Interpolation},
  journal      = {CoRR},
  volume       = {abs/2306.15595},
  year         = {2023},
  url          = {https://doi.org/10.48550/arXiv.2306.15595},
  doi          = {10.48550/ARXIV.2306.15595},
  eprinttype    = {arXiv},
  eprint       = {2306.15595},
  bibsource    = {dblp computer science bibliography, https://dblp.org}
}

@article{DBLP:journals/ijon/SuALPBL24,
  author       = {Jianlin Su and
                  Murtadha H. M. Ahmed and
                  Yu Lu and
                  Shengfeng Pan and
                  Wen Bo and
                  Yunfeng Liu},
  title        = {RoFormer: Enhanced transformer with Rotary Position Embedding},
  journal      = {Neurocomputing},
  volume       = {568},
  pages        = {127063},
  year         = {2024},
  url          = {https://doi.org/10.1016/j.neucom.2023.127063},
  doi          = {10.1016/J.NEUCOM.2023.127063},
  bibsource    = {dblp computer science bibliography, https://dblp.org}
}

@article{press2021train,
  title={Train short, test long: Attention with linear biases enables input length extrapolation},
  author={Press, Ofir and Smith, Noah A and Lewis, Mike},
  journal={arXiv preprint arXiv:2108.12409},
  year={2021}
}

@article{zhao2020seal,
  title={Seal: Segment-wise extractive-abstractive long-form text summarization},
  author={Zhao, Yao and Saleh, Mohammad and Liu, Peter J},
  journal={arXiv preprint arXiv:2006.10213},
  year={2020}
}

@article{zhang2021summ,
  title={Summ\^{} n: A multi-stage summarization framework for long input dialogues and documents},
  author={Zhang, Yusen and Ni, Ansong and Mao, Ziming and Wu, Chen Henry and Zhu, Chenguang and Deb, Budhaditya and Awadallah, Ahmed H and Radev, Dragomir and Zhang, Rui},
  journal={arXiv preprint arXiv:2110.10150},
  year={2021}
}

@inproceedings{cui2021sliding,
  title={Sliding selector network with dynamic memory for extractive summarization of long documents},
  author={Cui, Peng and Hu, Le},
  booktitle={Proceedings of the 2021 Conference of the North American Chapter of the Association for Computational Linguistics: Human Language Technologies},
  pages={5881--5891},
  year={2021}
}

@article{zaheer2020big,
  title={Big bird: Transformers for longer sequences},
  author={Zaheer, Manzil and Guruganesh, Guru and Dubey, Kumar Avinava and Ainslie, Joshua and Alberti, Chris and Ontanon, Santiago and Pham, Philip and Ravula, Anirudh and Wang, Qifan and Yang, Li and others},
  journal={Advances in neural information processing systems},
  volume={33},
  pages={17283--17297},
  year={2020}
}

@inproceedings{DBLP:conf/eacl/PangNKSZX23,
  author       = {Bo Pang and
                  Erik Nijkamp and
                  Wojciech Kryscinski and
                  Silvio Savarese and
                  Yingbo Zhou and
                  Caiming Xiong},
  editor       = {Andreas Vlachos and
                  Isabelle Augenstein},
  title        = {Long Document Summarization with Top-down and Bottom-up Inference},
  booktitle    = {Findings of the Association for Computational Linguistics: {EACL}
                  2023, Dubrovnik, Croatia, May 2-6, 2023},
  pages        = {1237--1254},
  publisher    = {Association for Computational Linguistics},
  year         = {2023},
  url          = {https://doi.org/10.18653/v1/2023.findings-eacl.94},
  doi          = {10.18653/V1/2023.FINDINGS-EACL.94},
  bibsource    = {dblp computer science bibliography, https://dblp.org}
}

@article{goyal2022news,
  title={News summarization and evaluation in the era of gpt-3},
  author={Goyal, Tanya and Li, Junyi Jessy and Durrett, Greg},
  journal={arXiv preprint arXiv:2209.12356},
  year={2022}
}

@article{pu2023summarization,
  title={Summarization is (almost) dead},
  author={Pu, Xiao and Gao, Mingqi and Wan, Xiaojun},
  journal={arXiv preprint arXiv:2309.09558},
  year={2023}
}

@article{keswani2024abstractive,
  title={Abstractive long text summarization using large language models},
  author={Keswani, Gunjan and Bisen, Wani and Padwad, Hirkani and Wankhedkar, Yash and Pandey, Sudhanshu and Soni, Ayushi},
  journal={International Journal of Intelligent Systems and Applications in Engineering},
  volume={12},
  number={12s},
  pages={160--168},
  year={2024}
}

@inproceedings{DBLP:conf/acl/RatnerLBRMAKSLS23,
  author       = {Nir Ratner and
                  Yoav Levine and
                  Yonatan Belinkov and
                  Ori Ram and
                  Inbal Magar and
                  Omri Abend and
                  Ehud Karpas and
                  Amnon Shashua and
                  Kevin Leyton{-}Brown and
                  Yoav Shoham},
  editor       = {Anna Rogers and
                  Jordan L. Boyd{-}Graber and
                  Naoaki Okazaki},
  title        = {Parallel Context Windows for Large Language Models},
  booktitle    = {Proceedings of the 61st Annual Meeting of the Association for Computational
                  Linguistics (Volume 1: Long Papers), {ACL} 2023, Toronto, Canada,
                  July 9-14, 2023},
  pages        = {6383--6402},
  publisher    = {Association for Computational Linguistics},
  year         = {2023},
  url          = {https://doi.org/10.18653/v1/2023.acl-long.352},
  doi          = {10.18653/V1/2023.ACL-LONG.352},
  bibsource    = {dblp computer science bibliography, https://dblp.org}
}

@inproceedings{DBLP:conf/acl/YenG024,
  author       = {Howard Yen and
                  Tianyu Gao and
                  Danqi Chen},
  editor       = {Lun{-}Wei Ku and
                  Andre Martins and
                  Vivek Srikumar},
  title        = {Long-Context Language Modeling with Parallel Context Encoding},
  booktitle    = {Proceedings of the 62nd Annual Meeting of the Association for Computational
                  Linguistics (Volume 1: Long Papers), {ACL} 2024, Bangkok, Thailand,
                  August 11-16, 2024},
  pages        = {2588--2610},
  publisher    = {Association for Computational Linguistics},
  year         = {2024},
  url          = {https://doi.org/10.18653/v1/2024.acl-long.142},
  doi          = {10.18653/V1/2024.ACL-LONG.142},
  bibsource    = {dblp computer science bibliography, https://dblp.org}
}

@article{DBLP:journals/corr/abs-2402-11550,
  author       = {Jun Zhao and
                  Can Zu and
                  Hao Xu and
                  Yi Lu and
                  Wei He and
                  Yiwen Ding and
                  Tao Gui and
                  Qi Zhang and
                  Xuanjing Huang},
  title        = {LongAgent: Scaling Language Models to 128k Context through Multi-Agent
                  Collaboration},
  journal      = {CoRR},
  volume       = {abs/2402.11550},
  year         = {2024},
  url          = {https://doi.org/10.48550/arXiv.2402.11550},
  doi          = {10.48550/ARXIV.2402.11550},
  eprinttype    = {arXiv},
  eprint       = {2402.11550},
  bibsource    = {dblp computer science bibliography, https://dblp.org}
}
\bibliographystyle{iclr2026_conference}

\clearpage
\appendix
\section{LLM-as-a-Judge Evaluation}
\label{appsec:llm-as-judge}

\subsection{LLM-as-a-Judge Setup}

We employ an LLM-as-a-Judge evaluation framework to assess the relative quality of generated summaries. For each document, we compare pairs of summaries using the evaluation prompt shown below. To mitigate positional bias, we reverse the order of summaries for each pair, ensuring that each comparison is evaluated twice with alternating positions. The judge is instructed to determine which summary better meets the evaluation criteria or whether they are of equal quality.

\begin{LLM-as-a-Judge}{Evaluation Prompt}{}
    $\mathcal{SYS \ PROMPT}$: 

    \vspace{0.5em}

    You are an expert evaluator tasked with objectively assessing the quality of text summarizations.
    
    Your response must strictly follow this format:

    Reasoning: [Brief, precise explanation based on the criteria above.]
    
    Better one or Equal: [Summary 1 or Summary 2 or Equal]

    \vspace{1em}
    \hrule\vspace{1pt}\hrule
    \vspace{1em}

    $\mathcal{USER \ PROMPT}$: 

    \vspace{0.5em}

    Evaluate the following document and two summaries provided below. Determine which summary better meets the evaluation criteria provided, or whether they are equal.

    Document:
    ``\{text\}''

    Summary 1:
    ``\{summary 1\}''

    Summary 2:
    ``\{summary 2\}''
    
\end{LLM-as-a-Judge} 

\subsection{Additional Results using \gptfourone as the Judge}

\begin{figure}[t]
\centering          
\subfigure[\oursensemble vs. \gptfouro]{\label{fig:judge-qs_4o_by_41}\includegraphics[width=0.48\linewidth]{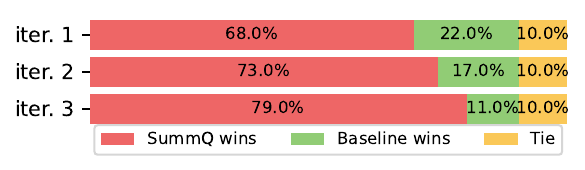}} 
\subfigure[\oursensemble vs. \gptothree]{\label{fig:judge-tool}\includegraphics[width=0.48\linewidth]{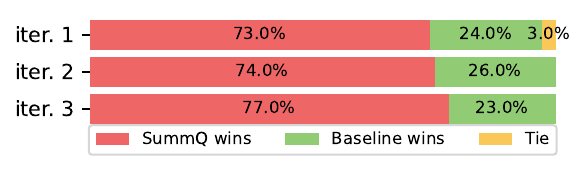}} 
\caption{The comparison between \oursensemble and baselines on \mensa  judged by \gptfourone during iteration, where there are three \gptfouro agents in each component of \oursensemble.}
\label{fig:appendix-llm-judge-evolve}
\end{figure}

We provide additional LLM-as-a-judge results judged by \gptfourone in \autoref{fig:appendix-llm-judge-evolve}, complementing the \gptfive results in \autoref{fig:llm-judge-evolve}. These results show that \oursensemble consistently outperforms all baselines across iterations, further validating the effectiveness of our method.

\section{Combination of different LLMs}
\label{appsec:combination}
\begin{table}[t]
\centering
\small
\setlength{\tabcolsep}{3.8pt}
\caption{
Results with different combinations of LLMs in each component on \mensa with the \oursensemble, \cmark\xspace indicates the LLM is used.}
\begin{tabular}{cccccccccc}
\toprule
\gptfouro & \gptfourone & \gptothree & \gptfive & \rougeone       & \rougetwo       & \rougel  &  \bsp & \bsr & \bsf \\ \midrule
\cmark \cmark \cmark  &   &  &   &  41.58   & 11.08   & 18.24   & 63.28   & 62.28   & 62.76  \\
\cmark  & \cmark    &      & \cmark    & 48.82 & \textbf{13.79} & 21.46 & 63.32 & 64.66 & 63.97 \\
\cmark  & \cmark & \cmark &  & 48.06 & 13.17 & 21.44 & 62.63 & 64.25 & 63.42 \\
\cmark  &  & \cmark & \cmark & \textbf{49.42}& 13.26& \textbf{22.90}& \textbf{63.71}& \textbf{65.17}& \textbf{64.42}\\
\bottomrule
\end{tabular}
\label{tab:multi_combination}
\end{table}

\paragraph{Diverse agent combinations leverage complementary strengths.}
We have demonstrated the effectiveness of \oursensemble with multiple \gptfouro agents, but we also explore the impact of combining different LLMs within our framework. \autoref{tab:multi_combination} presents results for various strategies that mix agent types, including \gptfouro, \gptfourone, \gptothree, and \gptfive. The results show that ensembles composed of diverse LLMs generally outperform those using a single model type. These combinations leverage complementary strengths, such as distinct reasoning capabilities, knowledge domains, or summarization styles. For instance, the pairing of \gptothree and \gptfive achieves the highest scores, likely due to \gptothree's advanced reasoning and \gptfive's robust routing and selection abilities. This diversity enables the system to generate more comprehensive and higher-quality summaries, as agents can mitigate each other's weaknesses and collectively address a broader range of content and quality challenges.

\section{Accuracy of Questions Evolve}
\label{appsec:acc-evolve}
\begin{wrapfigure}[16]{r}{5cm}
  \centering
    \includegraphics[scale=0.6]{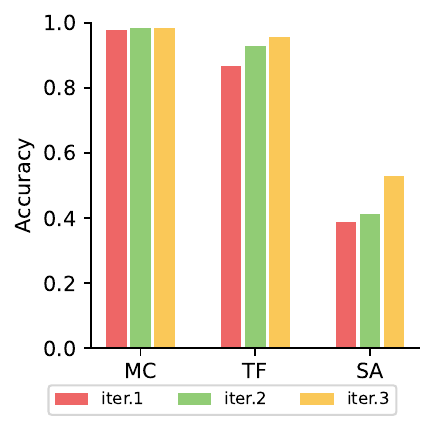}
    \caption{
    Accuracy of multiple-choice (MC), True-False (TF), and Short-Answer (SA) questions evolve during iteration on \mensa with the \oursensemble .
    }
    \label{fig:acc-evolve}
\end{wrapfigure}

\paragraph{Quiz answer accuracy improves throughout iteration.}
As the iteration proceeds, the accuracy of answering quiz questions based on the generated summaries improves steadily. \autoref{fig:acc-evolve} illustrates that all the question types exhibit this upward trend: multiple-choice (MC) and true-false (TF) questions achieve higher accuracy more rapidly, while short-answer (SA) questions, which demand deeper comprehension, show a more gradual improvement. This pattern underscores that agentic collaboration and iterative refinement in \ours enhance summary quality, as reflected in the improved performance on increasingly complex quiz questions.

\section{Cost Analysis}
\label{appsec:cost}

The cost analysis in \autoref{tab:cost} reveals significant differences in computational resource requirements across different approaches. While simple prompting baseline maintains the lowest cost at \$0.18 per instance with minimal input and output token usage, \ourssolo demonstrates a moderate increase in resource consumption, requiring \$1.95 per instance with 0.43M input tokens and 6.97K output tokens. This reflects the additional computational overhead of our iterative refinement process compared to baseline approaches. The \oursensemble variant shows the highest resource requirements at \$14.45 per instance, consuming 3.30M input tokens and generating 24.96K output tokens, which is attributable to the collaborative multi-agent framework involving multiple summary and quiz generators, reviewers, and the iterative debate process. Despite the higher computational cost, the substantial improvements in summary quality and quiz generation accuracy demonstrated throughout our evaluation justify this investment, particularly for applications where high-quality outputs are prioritized over computational efficiency.

\begin{table}[h]
\centering
\small
\caption{Average token usage and cost (in USD) per example of different methods on \mensa with \gptfouro.}
\begin{tabular}{lcccc}
\toprule
          & Input Tokens & Output Tokens & Cost (USD)  \\ \midrule
Prompting & 0.04M        & \phantom{0}0.23K         & \phantom{0}0.18  \\
\ourssolo     & 0.43M        & \phantom{0}6.97K         & \phantom{0}1.95  \\
\oursensemble       & 3.30M        & 24.96K        & 14.45 \\ \bottomrule
\end{tabular}
\label{tab:cost}
\end{table}

\section{Human Evaluation}
\label{appsec:human_eval}

We conduct a human evaluation to assess the quality of summaries generated by different methods. It is impractical for human judges to read the entire long documents from an unfamiliar domain, so we randomly select 20 NLP papers published after June 2024, and employ five Ph.D. students who published at least one NLP paper as judges. Each judge is presented with the source document and two summaries generated by different methods, and they are asked to choose the better summary considering the following aspects, including \textit{Informativeness}, \textit{Coherence}, and \textit{Factuality}. Based on the feedback from the judges that the summaries generated by \oursensemble (2084 words on average) are significantly longer than those from other methods (1450 words on average by \gptothree and 982 words on average by \gptfouro), we also include a rephrased version of \oursensemble, denoted as \oursensemblerephrase, where we prompt \gptfouro to shorten the summary generated by \oursensemble. The rephrased summaries have an average length of 1573 words, which is more comparable to the baselines. The selected papers are presented in \autoref{tab:paper}.

\begin{table}[t]
\centering
\small
\begin{tabular}{lp{11cm}}
\toprule
\textbf{arXiv ID} & \textbf{Title}\\ \midrule
2410.07095 & MLE-bench: Evaluating machine learning agents on machine learning engineering \\ 
2504.13959 & AI Safety should prioritize the Future of Work \\
2410.15522 & M-RewardBench: Evaluating Reward Models in Multilingual Settings \\
2406.17557 & The FineWeb Datasets: Decanting the Web for the Finest Text Data at Scale  \\
2411.01493  & Sample-Efficient Alignment for LLMs  \\
2407.19056 & Benchmarking Language Agents across Multiple Applications for Office Automation  \\
2411.19943  & Critical Tokens Matter: Token-Level Contrastive Estimation Enhances LLM's Reasoning Capability  \\
2412.03679 & Evaluating Language Models as Synthetic Data Generators  \\
2412.03555  & PaliGemma 2: A Family of Versatile VLMs for Transfer  \\
2412.09871 & Byte Latent Transformer: Patches Scale Better Than Tokens  \\
2412.06559 & ProcessBench: Identifying Process Errors in Mathematical Reasoning  \\
2412.14161 & TheAgentCompany: Benchmarking LLM Agents on Consequential Real World Tasks  \\
2406.06144 & Language Models Resist Alignment: Evidence From Data Compression  \\
2410.12883 & Scaling Laws for Multilingual Language Models  \\
2410.04840 & Strong Model Collapse  \\
2406.15480  & On Giant's Shoulders: Effortless Weak to Strong by Dynamic Logits Fusion  \\
2410.08964 & Language Imbalance Driven Rewarding for Multilingual Self-improving  \\
2411.19799 & INCLUDE: Evaluating Multilingual Language Understanding with Regional Knowledge  \\
2410.07825 & Extracting and Transferring Abilities For Building Multi-lingual Ability-enhanced Large Language Models  \\
2502.17910 & Scaling LLM Pre-training with Vocabulary Curriculum \\
\bottomrule
\end{tabular}
\caption{List of papers used for human evaluation, including arXiv ID and title.}
\label{tab:paper}
\end{table}

\begin{table}[t]
\centering
\small
\setlength{\tabcolsep}{7pt}
\begin{tabular}{p{0.1\textwidth}p{0.8\textwidth}}
\toprule
Model & Summary \\ \midrule
\gptfouro & The text describes various scenes and interactions from a screenplay set in South Boston and at M.I.T. It opens with a St. Patrick's Day parade and moves into a bar where four young men-Chuckie, Will, Billy, and Morgan-share a crude story. Will Hunting, a 20-year-old with remarkable intelligence, is revealed to be working as a janitor at M.I.T. and anonymously solving complex math problems on chalkboards. Professor Lambeau discovers Will's talents and arranges for him to avoid jail by meeting with therapists, including Sean Maguire, a tough psychologist from South Boston. Will's relationship with Skylar, a Harvard student, is explored, revealing his fear of intimacy and abandonment. Sean helps Will confront his past and emotional barriers. Will ultimately decides to pursue a job at Tri-tech, but leaves a note for Sean saying he had to go \"see about a girl,\" indicating he chose to follow Skylar to California. The story ends with Will driving away, symbolizing his decision to embrace change and new opportunities. \\ \midrule
\gptothree & Will Hunting, a 20-year-old orphan from South Boston, spends his days drinking with blue-collar friends and nights working as an MIT janitor. When he anonymously solves an impossible math proof, Professor Gerald Lambeau tracks him down, then saves him from jail (after an assault arrest) in exchange for weekly math work and therapy. Will sabotages several therapists until he meets Lambeau's former roommate, Sean Maguire, a community-college psychologist who grew up in the same neighborhood and is still mourning his late wife. Their confrontations force Will to face the childhood abuse, class anger and fear of abandonment that make him hide behind brilliance and bravado. While Lambeau lines up lucrative jobs, Will half-heartedly dates Harvard pre-med Skylar but pushes her away when she asks him to move to California. \textcolor{blue}{Chuckie, his best friend, challenges Will not to waste his ``lottery ticket.'' At last, a breakthrough with Sean convinces Will he deserves love and a future of his own choosing.} Rejecting the safe corporate offer, \textcolor{blue}{Will leaves Boston-after gifting his therapist a note that reads, ``Sorry, I had to go see about a girl''}-and drives west to reunite with Skylar, finally ready to risk both his talent and his heart. \\ \midrule
\oursensemble & The screenplay ``Good Will Hunting'' follows Will Hunting, a young man from South Boston with a troubled past and extraordinary mathematical abilities. \textcolor{red}{Will spends his time with his close-knit group of friends, including Chuckie, Morgan, and Billy, in their working-class Irish neighborhood.} Despite his genius, Will works as a janitor at MIT, where he anonymously solves complex mathematical problems posted by Professor Lambeau, a Field's Medal winner. \textcolor{red}{The story begins with scenes from the South Boston St. Patrick's Day Parade and a humorous yet dark tale from Chuckie about his cousin Mikey Sullivan's misadventure with an injured cat. Will's life takes a significant turn after he gets into a fight with Bobby Champa, who hurts a woman with a bottle, and faces jail time.} Professor Lambeau intervenes, offering Will a deal to avoid prison if he agrees to study mathematics and attend therapy sessions. Will reluctantly starts therapy with several psychologists but only begins to open up with Sean Maguire, a therapist with a similar background who has faced personal hardships, including the loss of his wife to cancer and his military service. Through their sessions, Sean helps Will confront his past traumas and fears of abandonment. Will's relationship with Skylar, a Harvard student, becomes strained as he struggles with his emotions and the idea of leaving his familiar life. \textcolor{red}{Despite being offered a job at the National Security Agency, Will refuses, citing ethical concerns. He initially considers a position at Tri-Tech Laboratories} but ultimately decides to follow his heart and go after Skylar, who has moved to California. The story concludes with Will leaving his old life behind, symbolized by his friends discovering his departure and Sean finding a note from Will explaining his decision to seek a future with Skylar. \textcolor{red}{Chuckie gifts Will a car for his 21st birthday, which Will uses to drive away, leaving his friends and old life behind.} \\ \bottomrule

\end{tabular}
\caption{
Case study on the summaries given by \gptfouro, \gptothree, and \oursensemble about the movie ``Good Will Hunting''. \textcolor{red}{The text in red} highlights the details covered by \oursensemble but missed by either \gptfouro or \gptothree. \textcolor{blue}{The text in blue} highlights the details covered by either \gptfouro and \gptothree but missed by \oursensemble.
}
\label{tab:case_study}
\end{table}
\section{Case Study}
\label{sec:appendix_case_study}

We present a case study in \autoref{tab:case_study} comparing the summaries generated by \gptfouro, \gptothree, and \oursensemble for the movie ``Good Will Hunting''. The text highlighted in red indicates details that are covered by \oursensemble but missed by either \gptfouro or \gptothree. Conversely, the text highlighted in blue represents details that are included in either \gptfouro or \gptothree but omitted by \oursensemble. This comparison illustrates how \oursensemble captures a broader range of important plot points and character developments, while also highlighting some specific elements that were overlooked. Overall, the case study demonstrates the strengths of our approach in generating comprehensive and detailed summaries.

\clearpage
\section{Prompts}
\label{appsec:prompts}
\subsection{Prompts of Summary Generators}

In this section, we present the detailed prompts used for summary generation in \oursensemble, covering draft summary generation, refinement, aggregation, best summary selection, and voting.

\begin{SummGen}{Draft Summary Generation Prompt}{}
    $\mathcal{SYS \ PROMPT}$: 

    \vspace{0.5em}
    
    You are a helpful assistant tasked with summarizing long text. Summarize the following text concisely and accurately, ensuring that all key points are covered. The summary should be clear and coherent, avoiding unnecessary details or repetition. Use precise language and maintain the original meaning of the text.

    \vspace{1em}
    \hrule\vspace{1pt}\hrule
    \vspace{1em}

    $\mathcal{USER \ PROMPT}$: 

    \vspace{0.5em}

    Original Text:
    ``\{Document\}''\\

    Summary:
    
\end{SummGen}

\begin{SummGen}{Refine Summary Generation Prompt}{}
    $\mathcal{SYS \ PROMPT}$: 

    \vspace{0.5em}
    
    You are a helpful assistant tasked with refining summaries. Given the original text, the initial summary, feedback from the evaluator, and feedback from quiz testing, refine the summary to better capture the key points in the original text and address any shortcomings.

    \vspace{1em}
    \hrule\vspace{1pt}\hrule
    \vspace{1em}

    $\mathcal{USER \ PROMPT}$: 

    \vspace{0.5em}

    Original Text:
     ``\{Document\}''

    Previous Summary:
    ``\{Summary\}''

    Reviewers Feedback:
    ``\{Summary reviewers feedback\}''

    Quiz Testing Feedback:
    
    The summary could not answer the following questions correctly:
    ``\{Examinee feedback\}''\\

    Refined Summary:
    
\end{SummGen}

\begin{SummGen}{Summary Aggregation Prompt}{}
    $\mathcal{SYS \ PROMPT}$: 

    \vspace{0.5em}
    
    You are an expert synthesiser. You will be given several candidate summaries of the same original text. Merge them into ONE superior summary that retains every important detail but avoids redundancy.

    \vspace{1em}
    \hrule\vspace{1pt}\hrule
    \vspace{1em}

    $\mathcal{USER \ PROMPT}$: 

    \vspace{0.5em}

    Original Text:
    ``\{Document\}''

    Candidate Summaries:
    ``\{Candidates\}''\\

    Merged Summary:
    
\end{SummGen} 

\begin{SummGen}{Best Summary Selection Prompt}{}
    $\mathcal{SYS \ PROMPT}$: 

    \vspace{0.5em}
    
    You are an expert summarisation judge. Rank the candidate summaries from best to worst according to coverage, factual accuracy and conciseness. Return the best summary.

    \vspace{1em}
    \hrule\vspace{1pt}\hrule
    \vspace{1em}

    $\mathcal{USER \ PROMPT}$: 

    \vspace{0.5em}

    Original Text:
    ``\{Document\}''

    Candidate Summaries:
    ``\{Candidates\}''\\

    Best Summary:
    
\end{SummGen} 

\begin{SummGen}{Voting Prompt}{}
    $\mathcal{SYS \ PROMPT}$: 

    \vspace{0.5em}
    
    You are an expert and strict summarization judge. Given two summaries, determine which one is better according to coverage, factual accuracy and conciseness. ONLY Return 1 or 2, where 1 means the first summary is better and 2 means the second summary is better. If both are equally good, return 1 or 2. Reply with nothing else.

    \vspace{1em}
    \hrule\vspace{1pt}\hrule
    \vspace{1em}

    $\mathcal{USER \ PROMPT}$: 

    \vspace{0.5em}

    Original Text:
    ``\{Document\}''

    Candidate Summaries:
    ``\{Candidates\}''\\

    Best One (1 or 2):
    
\end{SummGen}

\subsection{Prompts of Quiz Generators}

In this section, we present the detailed prompts for quiz generation in \oursensemble, covering draft quiz generation, refinement, aggregation, best quiz selection, and voting.

\begin{QuizGen}{Draft Quiz Generation Prompt}{}
    $\mathcal{SYS \ PROMPT}$: 

    \vspace{0.5em}

    Multiple-choice questions:
    
    Format:
    
    1. Question?
    
    A) Option 1
    
    B) Option 2
    
    C) Option 3
    
    D) Option 4 \\

    True/False questions:
    
    Format:
    
    1. Statement. (True/False)

    Short-answer question: \\
    
    Format:
    
    1. Question? \\

    You are a helpful assistant tasked with generating questions from long text. Generate quiz questions clearly covering key points. Include: ``\{num of mc\}'' Multiple-choice questions, ``\{num of tf\}'' True/False questions, and ``\{num of sa\}'' Short-answer question.. The Question Format is as above.

    \vspace{1em}
    \hrule\vspace{1pt}\hrule
    \vspace{1em}

    $\mathcal{USER \ PROMPT}$: 

    \vspace{0.5em}

    Original Text:
    ``\{Document\}''\\

    Quiz:
    
\end{QuizGen}

\begin{QuizGen}{Refine Quiz Generation Prompt}{}
    $\mathcal{SYS \ PROMPT}$: 

    \vspace{0.5em}
    
    You are a helpful assistant tasked with refining generated questions. Given the text, the initial generated questions, feedback from the evaluator, and feedback from quiz testing, refine the questions to ensure they cover important information clearly and avoid trivial or overly detailed content. Return ``\{num of mc\}'' Multiple-choice questions, ``\{num of tf\}'' True/False questions, and ``\{num of sa\}'' Short-answer question.

    \vspace{1em}
    \hrule\vspace{1pt}\hrule
    \vspace{1em}

    $\mathcal{USER \ PROMPT}$: 

    \vspace{0.5em}

    Original Text:
     ``\{Document\}''

    Previous Quiz:
    ``\{Quiz\}''

    Reviewers Feedback:
    ``\{Quiz reviewers feedback\}''

    Quiz Testing Feedback:
    
    The following questions could not be answered correctly based on the key information:
    ``\{Examinee feedback\}''\\

    Refined Quiz:
    
\end{QuizGen}

\begin{QuizGen}{Quiz Aggregation Prompt}{}
    $\mathcal{SYS \ PROMPT}$: 

    \vspace{0.5em}
    
    You are an expert synthesiser. You will be given several candidate generated questions of the same text. Merge them into superior questions that retains every important detail but avoids redundancy with ``\{num of mc\}'' Multiple-choice questions, ``\{num of tf\}'' True/False questions, and ``\{num of sa\}'' Short-answer question.

    \vspace{1em}
    \hrule\vspace{1pt}\hrule
    \vspace{1em}

    $\mathcal{USER \ PROMPT}$: 

    \vspace{0.5em}

    Original Text:
    ``\{Document\}''

    Candidate Quiz Sets:
    ``\{Candidates\}''\\

    Merged Quiz:
    
\end{QuizGen} 

\begin{QuizGen}{Best Quiz Selection Prompt}{}
    $\mathcal{SYS \ PROMPT}$: 

    \vspace{0.5em}
    
    You are an expert question generation judge. Rank the candidate questions sets from best to worst according to coverage, difficulty and clarity. Return the best question set.

    \vspace{1em}
    \hrule\vspace{1pt}\hrule
    \vspace{1em}

    $\mathcal{USER \ PROMPT}$: 

    \vspace{0.5em}

    Original Text:
    ``\{Document\}''

    Candidate Quiz Sets:
    ``\{Candidates\}''\\

    Best Quiz:
    
\end{QuizGen} 

\begin{QuizGen}{Voting Prompt}{}
    $\mathcal{SYS \ PROMPT}$: 

    \vspace{0.5em}
    
    You are an expert and strict question generation judge. Given two question sets, determine which one is better according to coverage, difficulty and clarity. ONLY Return 1 or 2, where 1 means the first question set is better and 2 means the second question set is better. If both are equally good, return 1 or 2. Reply with nothing else.

    \vspace{1em}
    \hrule\vspace{1pt}\hrule
    \vspace{1em}

    $\mathcal{USER \ PROMPT}$: 

    \vspace{0.5em}

    Original Text:
    ``\{Document\}''

    Candidate Quiz Sets:
    ``\{Candidates\}''\\

    Best One (1 or 2):
    
\end{QuizGen}

\subsection{Prompts of Summary Reviewers}

In this section, we provide the detailed prompts for summary review in \oursensemble, including summary review annotation, merging agreed issues, and debating contested issues.

\begin{SummReview}{Annotate Summary Prompt}{}
    $\mathcal{SYS \ PROMPT}$: 

    \vspace{0.5em}
    
    You are a strict generated summary reviewer. \\
    
    1. Coverage  - at least 90\% of key facts needed to answer every quiz question appear.
    
    2. Faithful  - no hallucinations / contradictions.
    
    3. Brevity   - $\leq$ 25 \% tokens of source OR $\leq$ 500 words.
    
    4. Clarity   - precise, coherent language. \\
    
    If **all four** points are satisfied output exactly `PASS' and reply with nothing else.
    
    Otherwise list concrete problems.
    
    For every problem output ONE line in the form:
    
    - CATEGORY: short description
    
    where CATEGORY is in \{COVERAGE, FAITHFUL, BREVITY, CLARITY\}.
    
    If there is no problem with this category, do not output this category.

    \vspace{1em}
    \hrule\vspace{1pt}\hrule
    \vspace{1em}

    $\mathcal{USER \ PROMPT}$: 

    \vspace{0.5em}

    Original Text:
    ``\{Document\}''

    Quiz Questions:
    ``\{questions\}''

    Summary to Review:
    ``\{summary\}''\\

    Feedback:
    
\end{SummReview}

\begin{SummReview}{Agreed Issues Merged Prompt}{}
    $\mathcal{SYS \ PROMPT}$: 

    \vspace{0.5em}
    
    You are an expert synthesiser. You will be given several feedback for the generated summary. Merge them into ONE superior feedback that retains every important detail but avoids redundancy.

    \vspace{1em}
    \hrule\vspace{1pt}\hrule
    \vspace{1em}

    $\mathcal{USER \ PROMPT}$: 

    \vspace{0.5em}

    Original Text:
    ``\{Document\}''

    Summary:
    ``\{Summary\}''

    Candidate Feedback:
    ``\{Candidates\}''\\

    Merged Feedback:
    
\end{SummReview}

\begin{SummReview}{Contested Issues Debate Prompt}{}
    $\mathcal{SYS \ PROMPT}$: 

    \vspace{0.5em}
    
    You are participating in a one-turn debate about the following alleged issue in a generated summary. Reply with ONE line starting with either KEEP (keep the issue) or DISCARD (discard the issue) followed by a brief justification.

    \vspace{1em}
    \hrule\vspace{1pt}\hrule
    \vspace{1em}

    $\mathcal{USER \ PROMPT}$: 

    \vspace{0.5em}

    Original Text:
    ``\{Document\}''
    
    Quiz Questions:
    ``\{questions\}''
    
    Summary:
    ``\{Summary\}''

    Issues to Debate:
    ``\{Issues\}''\\

    Feedback:
    
\end{SummReview}

\subsection{Prompts of Quiz Reviewers}

In this section, we provide the detailed prompts for quiz review in \oursensemble, including quiz review annotation, merging agreed issues, and debating contested issues.

\begin{QuizReview}{Annotate Quiz Prompt}{}
    $\mathcal{SYS \ PROMPT}$: 

    \vspace{0.5em}
    
    You are a strict question reviewer. 
    
    QUESTION-review rubric:\\

    A. Coverage Distribution  
    
    1. Every *major* section / scene / argument of the chapter is addressed.  
    
    2. No cluster: questions are spread across the beginning, middle, end. \\

    B. Cognitive Depth 
    
    • $\geq$ 40 \% Remember / Understand  
    
    • $\leq$ 20 \% Evaluate / Create  \\

    C. Format Balance  
    
    - Required counts of MC, True/False, Short-answer are respected.  
    
    - Short-answer asks for 1-2 sentences, names, dates, or concepts.  
    
    - MC: exactly 4 options, one correct; distractors plausible and non-overlapping. 
    
    - True/False: clear factual statements, no double-negatives. \\

    D. Difficulty Gradient  
    
    • 30 \% easy, 50 \% medium, 20 \% hard.  
        
    - Easy  : answer is stated explicitly.  
        
    - Medium: answer needs light inference / synthesis.  
    
    - Hard  : answer needs multi-sentence reasoning. \\

    E. Clarity \& Quality  
    
    1. Questions are grammatically correct, unambiguous, no trivia.  
    
    2. Each question targets *one* idea only.  
    
    3. No repeated facts across different questions.\\

    If **all** points are satisfied output exactly 'PASS' and reply with nothing else.
    
    Otherwise list concrete problems.
    
    For every problem output ONE line in the form:
    
    - CATEGORY: short description
    
    where CATEGORY is in \{Coverage Distribution, Cognitive Depth, Format Balance, Difficulty Gradient, Clarity \& Quality \}.
    
    If there is no problem with this category, do not output it.

    \vspace{1em}
    \hrule\vspace{1pt}\hrule
    \vspace{1em}

    $\mathcal{USER \ PROMPT}$: 

    \vspace{0.5em}

    Original Text:
    ``\{Document\}''

    Key Information:
    ``\{summary\}''
    
    Quiz to Review:
    ``\{Quiz\}''\\

    Feedback:
    
\end{QuizReview}

\begin{QuizReview}{Agreed Issues Merged Prompt}{}
    $\mathcal{SYS \ PROMPT}$: 

    \vspace{0.5em}
    
    You are an expert synthesiser. You will be given several feedback for the generated questions. Merge them into ONE superior feedback that retains every important detail but avoids redundancy.

    \vspace{1em}
    \hrule\vspace{1pt}\hrule
    \vspace{1em}

    $\mathcal{USER \ PROMPT}$: 

    \vspace{0.5em}

    Original Text:
    ``\{Document\}''

    Quiz:
    ``\{Quiz\}''

    Candidate Feedback:
    ``\{Candidates\}''\\

    Merged Feedback:
    
\end{QuizReview}

\begin{QuizReview}{Contested Issues Debate Prompt}{}
    $\mathcal{SYS \ PROMPT}$: 

    \vspace{0.5em}
    
    You are participating in a one-turn debate about the following alleged issue in the generated questions. Reply with ONE line starting with either KEEP (keep the issue) or DISCARD (discard the issue) followed by a brief justification.

    \vspace{1em}
    \hrule\vspace{1pt}\hrule
    \vspace{1em}

    $\mathcal{USER \ PROMPT}$: 

    \vspace{0.5em}

    Original Text:
    ``\{Document\}''
    
    Key Information:
    ``\{Summary\}''
    
    Quiz:
    ``\{Quiz\}''

    Issues to Debate:
    ``\{Issues\}''\\

    Feedback:
    
\end{QuizReview}

\subsection{Prompts of Examinee}

In this section, we present the detailed prompt for the examinee module in \oursensemble, which is responsible for answering the generated quizzes based on the provided summaries.

\begin{Examinee}{Take Quiz Prompt}{}
    $\mathcal{SYS \ PROMPT}$: 

    \vspace{0.5em}

    For every question below select the answer **in the required format**:

    ─ Multiple-choice → return only the correct letter (A/B/C/D).  
    
    ─ True/False      → return only the word True or False. 
    
    ─ Short-answer    → return a short phrase or sentence taken verbatim from the text (no extra commentary).

    \vspace{1em}
    \hrule\vspace{1pt}\hrule
    \vspace{1em}

    $\mathcal{USER \ PROMPT}$: 

    \vspace{0.5em}

    Text:
    ``\{Summary\}''

    Questions:
    ``\{Quiz Questions\}''\\

    Return one answer per line in the same order.
    
\end{Examinee}

\end{document}